\pdfoutput=1

\documentclass[11pt]{article}
\usepackage{booktabs} 
\usepackage{graphicx} 
\usepackage{acl}

\usepackage{times}
\usepackage{latexsym}

\usepackage[T1]{fontenc}
\usepackage{graphicx}
\usepackage{subfigure}
\usepackage{CJK}
\usepackage{multirow}
\usepackage{array}
\usepackage{booktabs} 
\usepackage{color}
\usepackage{graphicx}
\usepackage{amsfonts}
\usepackage{subfigure}
\usepackage{float}
\usepackage[linesnumbered,ruled,vlined]{algorithm2e}
\usepackage{amsmath}
\usepackage{xcolor}
\usepackage{adjustbox}
\usepackage{tabu}
\usepackage{array}
\usepackage[utf8]{inputenc}
\newcommand{\bluecomment}[1]{\textcolor{blue}{\tcp*[f]{#1}}} 

\usepackage{microtype}

\usepackage{inconsolata}

\title{Self-Modifying State Modeling for Simultaneous Machine Translation}

\author{
    Donglei Yu\textsuperscript{1,2}, \ 
    Xiaomian Kang\textsuperscript{1,2}, \ 
    Yuchen Liu\textsuperscript{1,2}, \ 
    Yu Zhou\textsuperscript{1,3}\thanks{\ \ Corresponding Author},   \ 
    Chengqing Zong \textsuperscript{1,2} \\
    \textsuperscript{1}State Key Laboratory of Multimodal Artificial Intelligence Systems, \\
Institute of Automation, Chinese Academy of Sciences, Beijing, China\\
    \textsuperscript{2}School of Artificial Intelligence, University of Chinese Academy of Sciences, Beijing, China\\
    \textsuperscript{3} Fanyu AI Laboratory, Zhongke Fanyu Technology Co., Ltd, Beijing, China\\
    \texttt{\{yudonglei2021,kangxiaomian2014,yuchen.liu,yu.zhou,chengqing.zong\}@ia.ac.cn} \\
}
\begin{document}
\maketitle
\begin{abstract}
Simultaneous Machine Translation (SiMT) generates target outputs while receiving stream source inputs and requires a read/write policy to decide whether to wait for the next source token or generate a new target token, whose decisions form a \textit{decision path}. Existing SiMT methods, which learn the policy by exploring various decision paths in training, face inherent limitations. These methods not only fail to precisely optimize the policy due to the inability to accurately assess the individual impact of each decision on SiMT performance, but also cannot sufficiently explore all potential paths because of their vast number. Besides, building decision paths requires unidirectional encoders to simulate streaming source inputs, which impairs the translation quality of SiMT models. To solve these issues, we propose \textbf{S}elf-\textbf{M}odifying \textbf{S}tate \textbf{M}odeling (SM$^2$), a novel training paradigm for SiMT task. Without building decision paths, SM$^2$ individually optimizes decisions at each state during training. To precisely optimize the policy, SM$^2$ introduces Self-Modifying process to independently assess and adjust decisions at each state. For sufficient exploration, SM$^2$ proposes Prefix Sampling to efficiently traverse all potential states. Moreover, SM$^2$ ensures compatibility with bidirectional encoders, thus achieving higher translation quality. Experiments show that SM$^2$ outperforms strong baselines. Furthermore, SM$^2$ allows offline machine translation models to acquire SiMT ability with fine-tuning \footnote{Our source code is available at \url{https://github.com/EurekaForNLP/SM2}}.
\end{abstract}

\section{Introduction}
Simultaneous Machine Translation (SiMT) \citep{gu2017learning,ma2019stacl,zhang2020learning} outputs translation while receiving the streaming source sentence. Different from normal Offline Machine Translation (OMT) \citep{vaswani2017attention}, SiMT needs a suitable read/write policy to decide whether to wait for the coming source inputs (READ) or generate target tokens (WRITE).

\begin{figure}[t]
	\centering
	\small
        \subfigure[Training paradigm based on decision paths]{
            \includegraphics[width=0.45\textwidth]{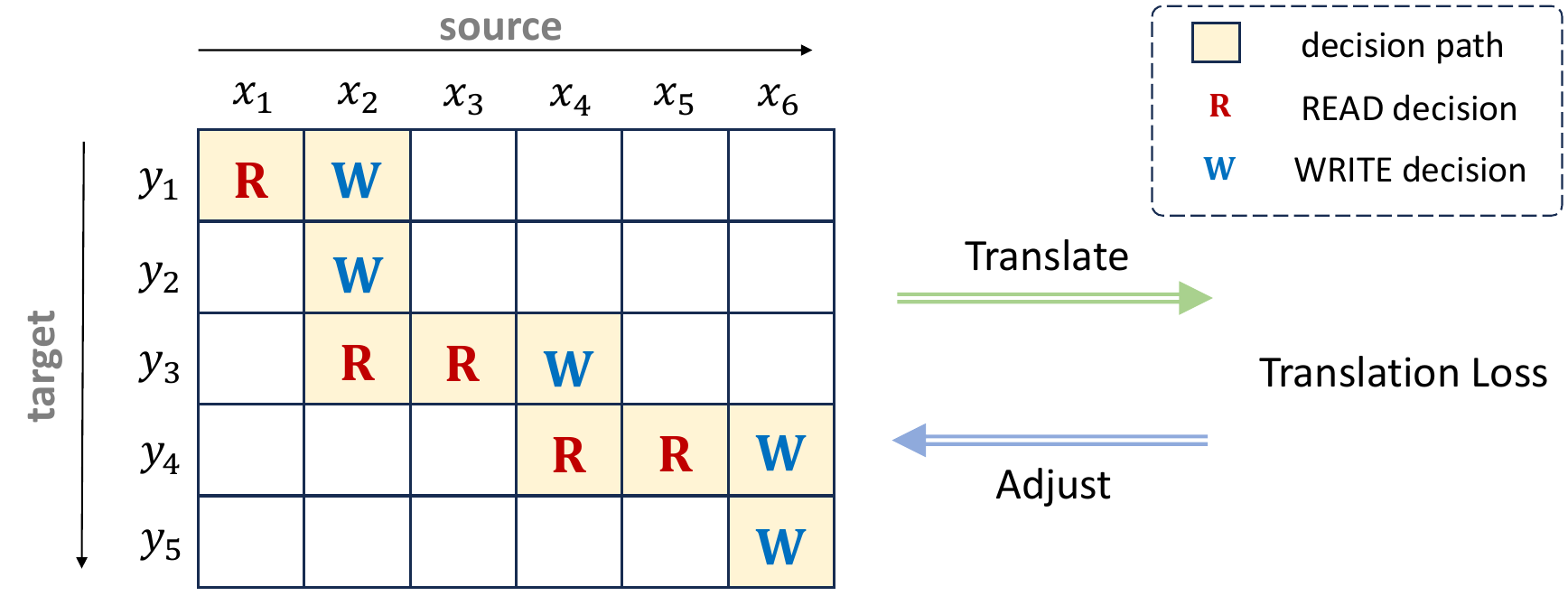} 
            \label{intro-case-ppl}
        }
        \subfigure[Our proposed Self-Modifying State Modeling]{
            \includegraphics[width=0.45\textwidth]{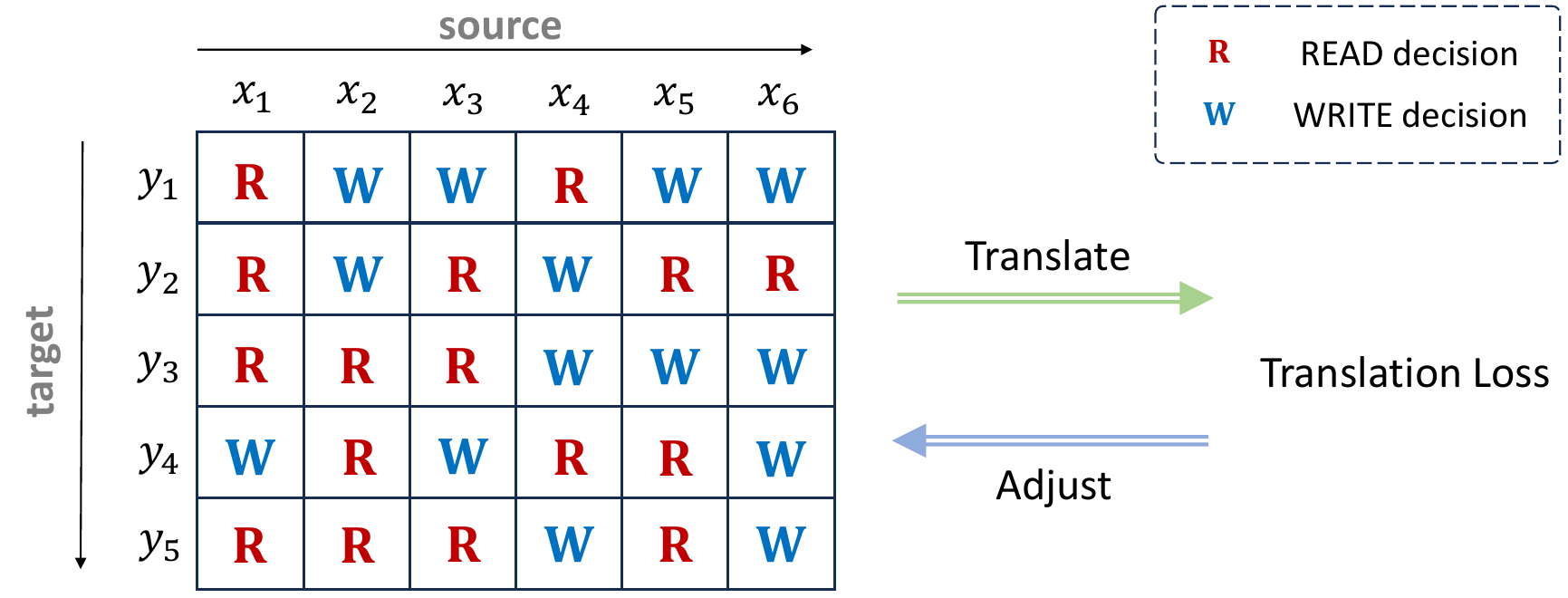}
        \label{intro-case-sm2}
        }
	\caption{Illustration of different paradigms. \textbf{(a)} Training paradigm based on decision paths. All decisions along a path are optimized in an integrated manner. \textbf{(b)} Self-Modifying State Modeling. The decisions at each state are optimized individually.}
	\label{intro-case}
    \end{figure}

As shown in Figure \ref{intro-case-ppl}, to learn a suitable policy, existing SiMT methods usually require building a \textit{decision path} (i.e., a series of READ and WRITE decisions made by the policy) to simulate the complete SiMT process during training \cite{zhang2022modeling}. Methods of fixed policies \citep{ma2019stacl,zhang2021universal} build the decision path based on pre-defined rules, and only optimize translation quality along the path. Methods of adaptive policies \citep{zheng2019simultaneous, miao2021generative, zhang2023hidden} dynamically build the decision path and optimize the policy based on the SiMT performance along this path.

However, the current training paradigm based on decision paths faces inherent limitations. First, it can lead to \textbf{imprecise optimization} of the policy during training. For fixed policies, pre-defined rules cannot ensure optimal decisions at each state. For adaptive policies, there exists a credit assignment problem \citep{minsky1961steps}, which means it is difficult to identify the impact of each individual decision on the global SiMT performance along a path, thus hindering the precise optimization of each decision. Second, due to numerous potential decision paths, existing methods \citep{zheng2019simultaneous, miao2021generative,zhang2023hidden} often prohibit the exploration of some paths during training, but this \textbf{insufficient exploration} cannot ensure the optimal policy. Third, for building decision paths in training, existing methods require \textbf{unidirectional encoders} to simulate streaming source inputs and avoid the leakage of source future information \citep{elbayad2020efficient}, which impairs SiMT models' translation quality \citep{iranzo2022simultaneous,kim2023enhanced}.

To address these issues, we propose \textbf{S}elf-\textbf{M}odifying \textbf{S}tate \textbf{M}odeling (SM$^2$), a novel training paradigm for SiMT task. As shown in Figure \ref{intro-case-sm2}, instead of constructing complete decision paths, SM$^2$ individually optimizes decisions at all potential states during training. This paradigm necessitates addressing two critical issues: firstly, how to independently optimize each decision based on its own contribution to SiMT performance; and secondly, how to sufficiently explore all potential states during training. To realize the independent optimization, SM$^2$ assesses each decision by estimating confidence values which measure the translation credibility. High confidence means the SiMT model can predict a credible target token at current state and WRITE is beneficial for SiMT performance; otherwise, READ is preferred. Since golden confidence values are unavailable, SM$^2$ introduces \textbf{Self-Modifying} process to learn accurate confidence estimation \citep{devries2018learning, lu2022learning}. Specifically, during training, the SiMT model is allowed to modify its prediction based on the received source prefix with the prediction based on the complete source sentence, and the confidence is estimated to determine whether the modification is necessary to ensure a credible prediction at current state. To sufficiently explore all potential states, SM$^2$ conducts \textbf{Prefix Sampling} to divide all states into groups according to the number of their received source prefix tokens, and sample one group for optimization in each iteration. 

Compared to the training paradigm based on decision paths, SM$^2$ presents significant advantages. First, the Self-Modifying process can assess each decision independently, which realizes the precise optimization of policy without the credit assignment problem. Second, Prefix Sampling ensures sufficient exploration of all potential states, promoting the discovery of the optimal policy. These benefits enable SM$^2$ to learn a more effective policy. Furthermore, without building decision paths in training, SM$^2$ ensures compatibility with bidirectional encoders, thereby improving translation quality. This compatibility also allows OMT models to acquire the SiMT capability via fine-tuning.
Our contributions are outlined in the following:
\begin{itemize}
    \item We propose \textbf{S}elf-\textbf{M}odifying \textbf{S}tate \textbf{M}odeling (SM$^2$), a novel training paradigm that individually optimizes decisions at all states without building complete decision paths.
    \item SM$^2$ can learn a better policy through precise optimization of each decision and sufficient exploration of all states. With bidirectional encoders, SM$^2$ achieves higher translation quality and compatibility with OMT models.
    \item Experimental results on Zh$\rightarrow$En, De$\rightarrow$En and En$\rightarrow$Ro SiMT tasks show that SM$^2$ outperforms strong baselines under all latency levels.
\end{itemize}

\section{Background}
\textbf{Simultaneous machine translation} For SiMT task, we respectively denote the source sentence as $\mathbf{x}=(x_1,...,x_M)$ and the corresponding target sentence as $\mathbf{y}=(y_1,..,y_N)$. Since the source inputs are streaming, we denote the number of source tokens available when generating $y_i$ as $g_i$, and hence the prediction probability of $y_i$ is $p (y_i \mid \mathbf{x}_{\leq g_i},\mathbf{y}_{< i})$ \citep{ma2019stacl}. Thus, the decoding probability of $\mathbf{y}$ is given by:
    \begin{equation}
        \begin{split}
           p\left ( \mathbf{y} \mid \mathbf{x} \right )=\prod_{i=1}^{N}p (y_i \mid \mathbf{x}_{\leq g_i},\mathbf{y}_{< i})
        \end{split}
        \label{prefix-to-prefix}
    \end{equation}

\begin{figure*}[t]
	\centering
	\small
        \includegraphics[width=1\textwidth]{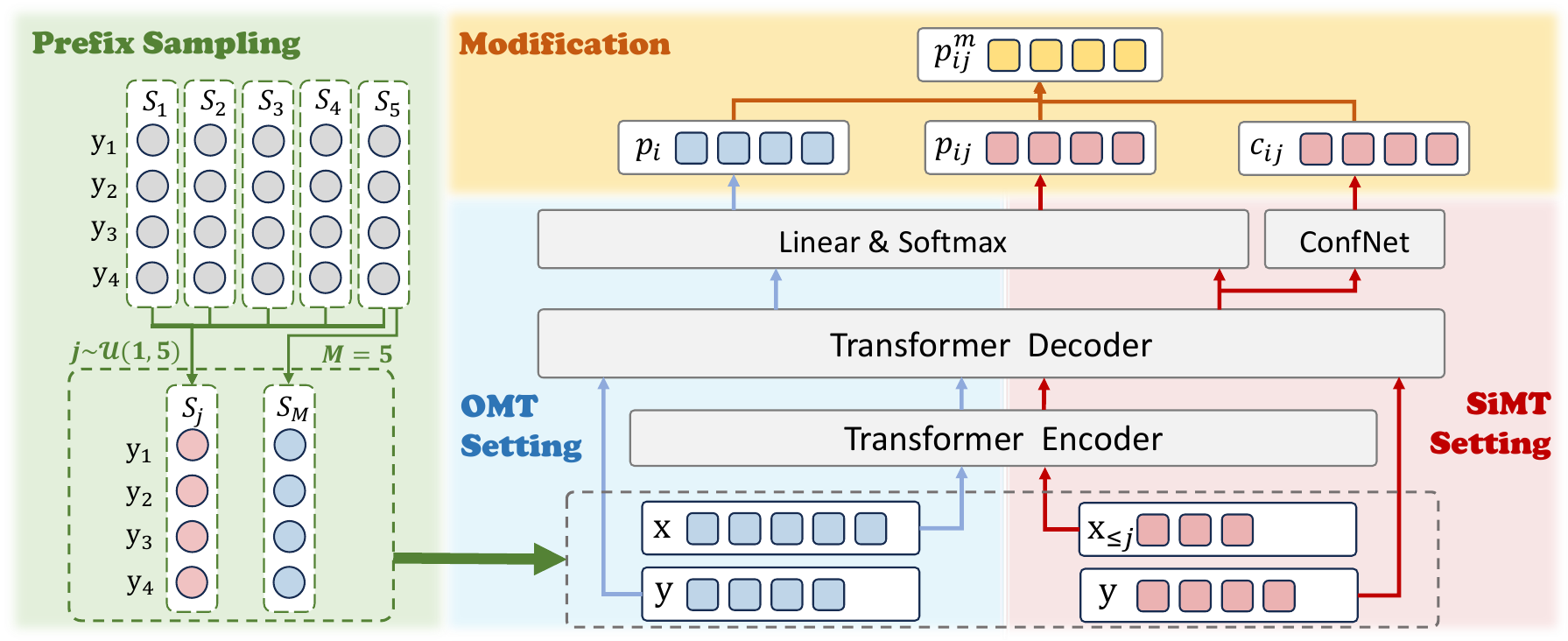}
	\caption{Overview of SM$^2$. $S_j$ contains the states where $\mathbf{x}_{\leq j}$ is received. $S_M$ contains the states where complete $\mathbf{x}$ is received. We introduce a confidence net (ConfNet) to estimate the confidence of each state. The model parameters in SiMT setting and OMT setting are shared. In this figure, the sentence lengths of the source and target sides are set to $M=5$ and $N=4$ respectively, and $j=3$ in the Prefix Sampling step.}
	\label{sm2-framework}
    \end{figure*}
\noindent\textbf{Decision state and decision path} We define the state $s_{ij}$ as the condition in which the source prefix $\mathbf{x}_{\leq j}$ has been received and the target prefix $\mathbf{y}_{<i}$ has been generated. At $s_{ij}$, a decision $d_{ij}\in \{\textrm{WRITE,READ}\}$ can be made based on the context $(\mathbf{x}_{\leq j},\mathbf{y}_{<i})$ \cite{grissom2014don,gu2017learning,zhao2023adaptive}. Specifically, if $\mathbf{x}_{\leq j}$ is sufficient for the SiMT model to predict $y_i$ accurately, $d_{ij}$ should be WRITE; otherwise, $d_{ij}$ should be READ. As shown in Figure \ref{intro-case-ppl}, a series of decisions $[d_{00},...,d_{NM}]$ are made in the SiMT process, which forms a decision path from $s_{00}$ to $s_{NM}$. Along the decision path, the SiMT model can finish reading the whole $\mathbf{x}$ and outputting the complete $\mathbf{y}$ \cite{zhang2022modeling}. Since these concepts are usually used in SiMT methods based on reinforcement learning (RL) \citep{grissom2014don,gu2017learning}, we compare our method with RL-based methods in Appendix \ref{comp-sm2-rl} for clarity.

\section{The Proposed Method}
We propose \textbf{S}elf-\textbf{M}odifying \textbf{S}tate \textbf{M}odeling (SM$^2$), which individually optimizes decisions at all states. The overview of SM$^2$ is shown in Figure \ref{sm2-framework}. To independently optimize each decision, SM$^2$ learns confidence estimation to assess decisions at each state by modeling the Self-Modifying process (Sec.\ref{self-modify-state-model}). To ensure sufficient exploration during training, SM$^2$ conducts Prefix Sampling to traverse all potential states (Sec. \ref{prefix-sampling}). Then, based on estimated confidence at each state, SM$^2$ can determine whether the received source tokens are sufficient to generate a credible token and make suitable decisions during inference (Sec.\ref{conf-based-infer}).
\subsection{Self-Modifying for Confidence Estimation}
\label{self-modify-state-model}

Intuitively, when a translation model has access to the complete input $\mathbf{x}$ (i.e., OMT setting), it can produce credible outputs. Therefore, a prediction made by the translation model at $s_{ij}$ (i.e. SiMT setting) is considered credible if it aligns with that in OMT setting. Conversely, if the prediction in SiMT setting is incredible, it will be modified in OMT setting. Based on this insight and \textit{Ask For Hints} \citep{devries2018learning, lu2022learning}, we model the Self-Modifying process to assess the translation credibility of each state. Specifically, we provide the SiMT model an option to modify its prediction in SiMT setting with that in OMT setting, and confidence estimation is defined as a binary classification determining whether the current generation requires the modification to ensure a credible prediction. Through measuring translation credibility, decisions at each state can be independently assessed. High confidence means the SiMT model can generate a credible token at $s_{ij}$ without modification and the WRITE decision is beneficial for SiMT performance; whereas low confidence indicates the prediction is inaccurate at $s_{ij}$ and the READ decision is preferred.

During training, the Self-Modifying process is conducted in two steps: \textit{prediction in SiMT setting \& OMT setting} and \textit{confidence-based modification}.

For \textit{prediction in SiMT setting\& OMT setting}, the SiMT model outputs different predictions at $s_{ij}$ in SiMT setting and OMT setting respectively. These predictions are calculated as follows: 
    \begin{equation}
        \begin{split}
            p_{ij}&=p(y_i\mid \mathbf{x}_{\leq j},\mathbf{y}_{< i})\\
            p_{i} &= p(y_i\mid\mathbf{x},\mathbf{y}_{< i})
        \end{split}
        \label{pred}
    \end{equation}
It is noted that the model parameters in SiMT setting and OMT setting are shared.

For \textit{confidence-based modification}, an additional confidence net is used to predict the confidence $c_{ij}$ at $s_{ij}$. The confidence net is represented as:
    \begin{equation}
        \begin{split}
            c_{ij}={\rm sigmoid}(W^T\cdot h_{ij}+b)
        \end{split}
        \label{conf-net}
    \end{equation}
where $h_{ij}$ is the hidden representation from the top decoder layer in SiMT setting and $\theta=\{W,b\}$ are trainable parameters. If $p_{ij}$ is credible, $c_{ij}$ should be close to 1; otherwise, $c_{ij}$ should be close to 0. To accurately calibrate $c_{ij}$ in the training process, we integrate the modification into the prediction probability as follows:
    \begin{equation}
        \begin{split}
           p^m_{ij}=c_{ij}\cdot p_{ij}+(1-c_{ij})\cdot p_i
        \end{split}
        \label{modified-prob}
    \end{equation}
Subsequently, the translation loss is calculated using the modified probability:
    \begin{equation}
        \begin{split}
           \mathcal{L}_{s_{ij}}=-y_i\log(p^m_{ij})
        \end{split}
        \label{modified-trans-loss}
    \end{equation}
Notably, the SiMT model can enhance the prediction credibility by estimating a lower $c_{ij}$ for more modification. However, this manner may cause an over-reliance on $p_i$. To avoid that, an additional penalty term for $c_{ij}$ is introduced:
    \begin{equation}
        \begin{split}
           \mathcal{L}_{c_{ij}}=-\log(c_{ij})
        \end{split}
        \label{penlty-loss}
    \end{equation}

Through Self-Modifying process, SM$^2$ independently optimizes each decision based on their individual effect on the SiMT performance, thus realizing the precise optimization of the policy without credit assignment problem. We provide a gradient analysis of the independent optimization in Appendix \ref{gradient-analysis} for further explanation.
\begin{algorithm}[t]
\caption{Confidence-based Policy}
\label{policy-algorithm}
\SetAlgoLined
\SetKwInOut{Input}{Input}\SetKwInOut{Output}{Output}
\Input{Streaming inputs $\mathbf{x}_{\leq j}$, Threshold $\gamma$, $i=1$, $j=1$, $y_0 \leftarrow \left \langle  \mathrm{BOS} \right \rangle$}
\Output{Target outputs $\mathbf{y}$}
\While{$y_{i-1} \neq \left \langle  \mathrm{EOS} \right \rangle$}{
    calculate confidence $c_{ij}$ as Eq.(\ref{conf-net})\;
    \eIf(\bluecomment{WRITE}){$c_{ij} \geq \gamma $}{
        generate $y_i$ with $\mathbf{x}_{\leq j},\mathbf{y}_{< i}$\;
        $i \leftarrow i+1$\;
        }(\bluecomment{READ}){
        wait for next source token $x_{j+1}$\;
        $j \leftarrow j+1$\;
    }
}
\end{algorithm}

\subsection{Prefix Sampling} 
\label{prefix-sampling}
To sufficiently explore all potential states during training, Prefix Sampling is conducted in SM$^2$. As shown in Figure \ref{sm2-framework}, states are categorized into groups, and one group is randomly sampled for optimization in each iteration. Specifically, all possible states of $(\mathbf{x},\mathbf{y})$ are divided into $M$ groups according to the number of their received source prefix tokens, and each group comprises $N$ states, which can be formulated as follows:
    \begin{equation}
        \begin{split}
           S_j=\{s_{ij}\mid 1 \leq i \leq N\}, j\in[1,M]
        \end{split}
        \label{state-groups}
    \end{equation}

In each iteration, we sample $j\sim{{\mathcal U}(1, M)}$. Then, SM$^2$ respectively predicts target translation in SiMT setting based on $S_j$ and those in OMT setting based on $S_{M}$, where the complete source sentence is received. Thus, the modified translation loss and the penalty item of each iteration are computed as follows:
    \begin{equation}
        \begin{split}
          \mathcal{L}_{S_j}&=\sum_{i=1}^{N}\mathcal{L}_{s_{ij}} \\
          \mathcal{L}_{C_j}&=\sum_{i=1}^{N}\mathcal{L}_{c_{ij}}
        \end{split}
        \label{group-trans-penalty-loss}
    \end{equation}

Besides, to ensure the $p_{i}$ in OMT setting can provide effective modification, the translation loss in OMT setting is required, which is formulated as:
    \begin{equation}
        \begin{split}
          \mathcal{L}_{omt}&=-\sum_{i=1}^N\log(p_i)
        \end{split}
        \label{group-omt-trans-loss}
    \end{equation}

The total training loss is the following:
    \begin{equation}
        \begin{split}
          \mathcal{L}=\mathcal{L}_{omt}+\mathcal{L}_{S_j}+\lambda\mathcal{L}_{C_j}
        \end{split}
        \label{group-loss}
    \end{equation}
where $\lambda$ is the super parameter. We discuss the effect of $\lambda$ in Appendix \ref{effect-of-lambda}.

Through Prefix Sampling, SM$^2$ explores all potential states without building any decision paths. Therefore, SM$^2$ can employ bidirectional encoders without the leakage of source future information in the training process.
\subsection{Confidence-based Policy in Inference}
\label{conf-based-infer}
During inference, SM$^2$ utilizes $c_{ij}$ to assess the credibility of current prediction, thus making suitable decisions between READ and WRITE at $s_{ij}$. Specifically, a confidence threshold $\gamma$ is introduced to serve as a criterion for making decisions. As shown in Algorithm \ref{policy-algorithm}, if $c_{ij} > \gamma$, SM$^2$ selects WRITE; otherwise, SM$^2$ selects READ. This decision process is constantly repeated until the complete translation is finished. It is noted that we only utilize SiMT setting in the inference process.

By adjusting $\gamma$, SM$^2$ can perform the SiMT task under different latency levels. A higher $\gamma$ encourages the SiMT model to predict more credible target tokens and the latency will be longer. Conversely, a lower $\gamma$ reduces the latency but may lead to a decrease in translation quality. The values of $\gamma$ employed in our subsequent experiments are detailed in Appendix \ref{hyper}.
\section{Experiments}
\subsection{Datasets}
We conduct experiments on three datasets:

\textbf{Zh$\rightarrow$En} We use LDC corpus which contains 2.1M sentence pairs as the training set, NIST 2008  for the validation set and NIST 2003, 2004, 2005, and 2006 for the test sets.

\textbf{De$\rightarrow$En} We choose WMT15 for training, which contains 4.5M sentence pairs. Newstest 2013 are used as the validation set and newstest 2015 are used as the test set.

\textbf{En$\rightarrow$Ro} WMT16 (0.6M) is used as the training set. We choose newsdev 2016 as the validation set and newstest 2016 as the test test.

We apply BPE \cite{sennrich2016neural} for all language pairs. In Zh$\rightarrow$En, the vocabulary size is 30k for Chinese and 20k for English. In both De$\rightarrow$En and En$\rightarrow$Ro, a shared vocabulary is learned with 32k merge operations. Additional experiments on WMT15 En$\rightarrow$Vi are provided in Appendix \ref{num-results}.
\subsection{System Settings}

The models used in our experiments are introduced as follows. All baselines are built based on Transformer \cite{vaswani2017attention} with the unidirectional encoder unless otherwise stated. More details are presented in Appendix \ref{hyper}.

    \textbf{OMT-Uni/OMT-Bi}\citep{vaswani2017attention}: OMT model with an unidirectional/bidirectional encoder.
    
    \textbf{wait-$k$} \citep{ma2019stacl}: a fixed policy, which first reads $k$ tokens, then writes one token and reads one token in turns.
    
    \textbf{m-wait-$k$} \citep{elbayad2020efficient}: a fix policy, which improves wait-$k$ by randomly sampling different $k$ during training.
    
    \textbf{ITST} \cite{zhang-feng-2022-information}: an adaptive policy, which models the SiMT task as a transport problem of information from source to target.
    
    \textbf{HMT} \citep{zhang2023hidden}: an adaptive policy, which models the SiMT task as a hidden Markov model, by treating the states as hidden events and the predicted tokens as observed events.
    
    \textbf{SM$^2$-Uni/SM$^2$-Bi}: Our proposed method with an unidirectional/bidirectional encoder.

\subsection{Evaluation Metric}
For SiMT, both translation quality and latency require evaluation. Since existing datasets mainly focus on the OMT task, the metric based on n-gram may cause inaccurate evaluation \citep{rei2020comet}. Therefore, we measure the translation quality with both SacreBLEU \citep{post2018call} and COMET\footnote{Unbabel/wmt22-cometkiwi-da} scores. For latency evaluation, we choose Average Lagging (AL) \cite{ma2019stacl} as the metric.

Furthermore, to assess the quality of read/write policy in different SiMT models, we follow \citet{zhang-feng-2022-information} and \citet{kim2023enhanced} to use Satisfied Alignments (SA), the proportion of the ground-truth aligned source tokens received before translating. Specifically, when generating $y_i$, the number of received source tokens $g_i$ should be no less than the golden-truth aligned source position $a_i$, so that the alignment between $y_i$ and $x_{a_i}$ can be satisfied in the SiMT process. Thus, SA($\uparrow$) can be calculated as:
    \begin{equation}
        \begin{split}
        {\rm SA} = \frac{1}{N}\sum_{i=1}^{N}\mathbb{I}(a_i\leq g_i)
        \end{split}
        \label{alignments-are-read}
    \end{equation}

\begin{figure*}[t]
	\centering
	\small
        \subfigure[Zh$\rightarrow$En]{
            \includegraphics[width=0.3\textwidth]{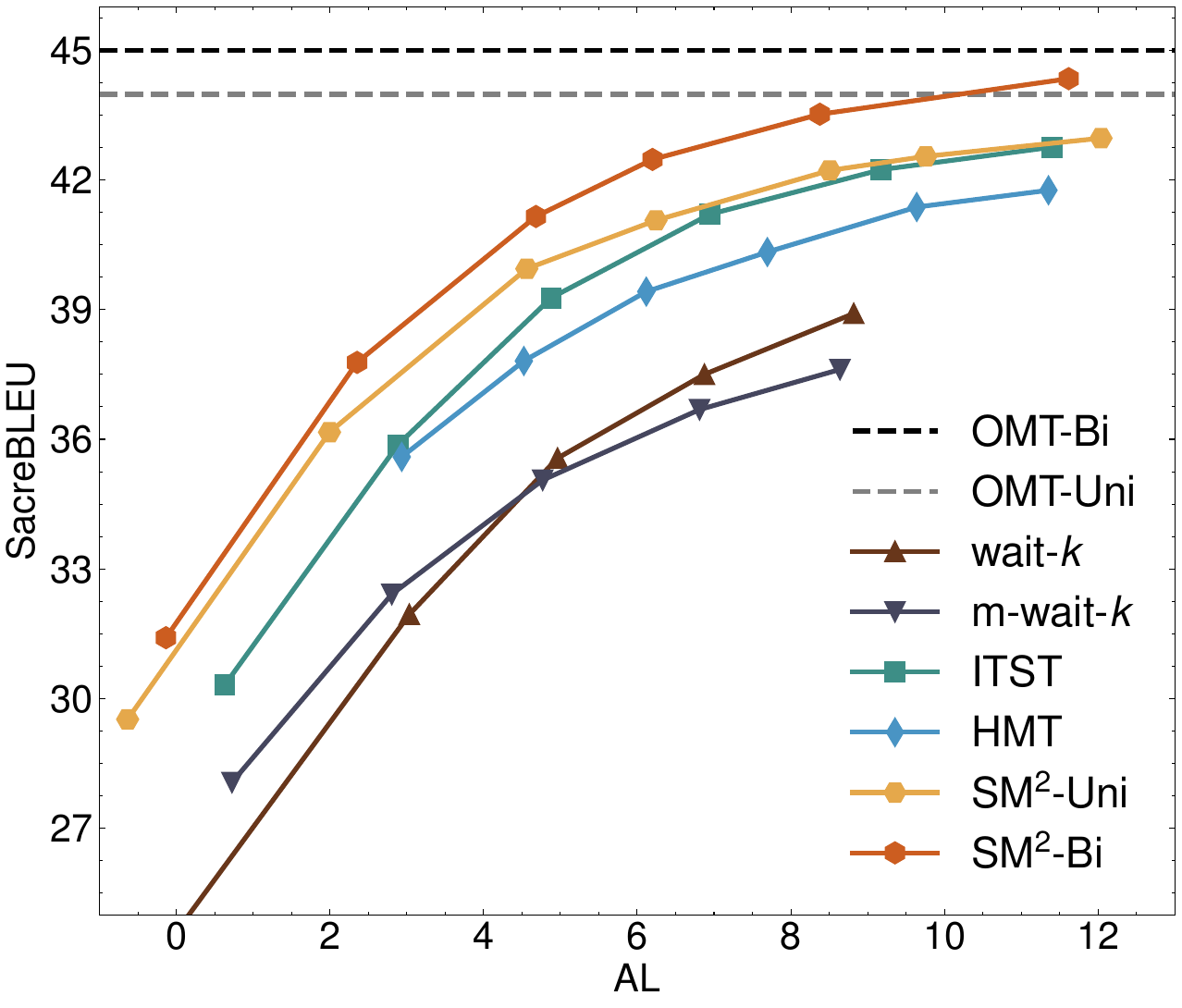} 
        }
        \subfigure[De$\rightarrow$En]{
            \includegraphics[width=0.3\textwidth]{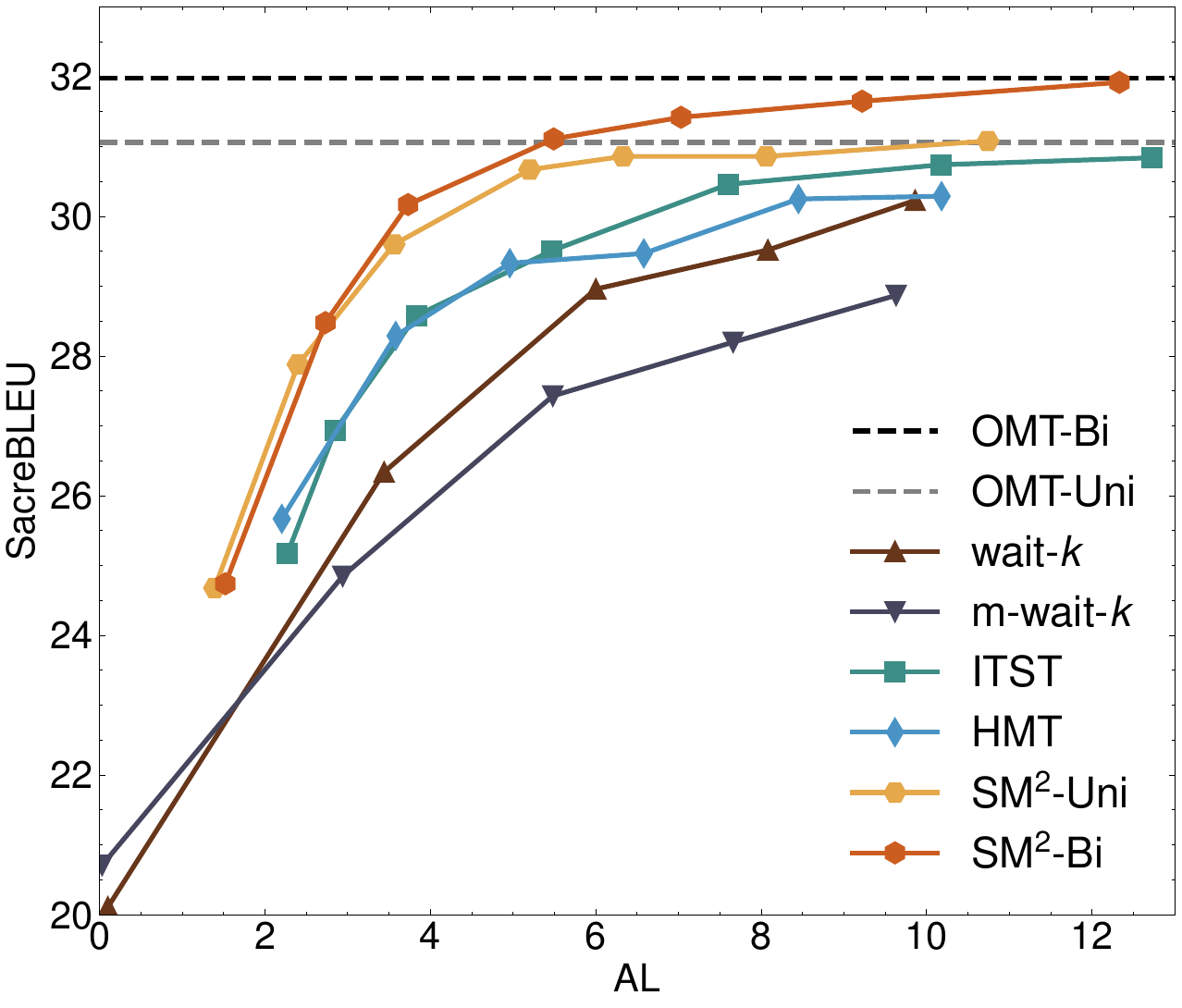}
        }
        \subfigure[En$\rightarrow$Ro]{
            \includegraphics[width=0.3\textwidth]{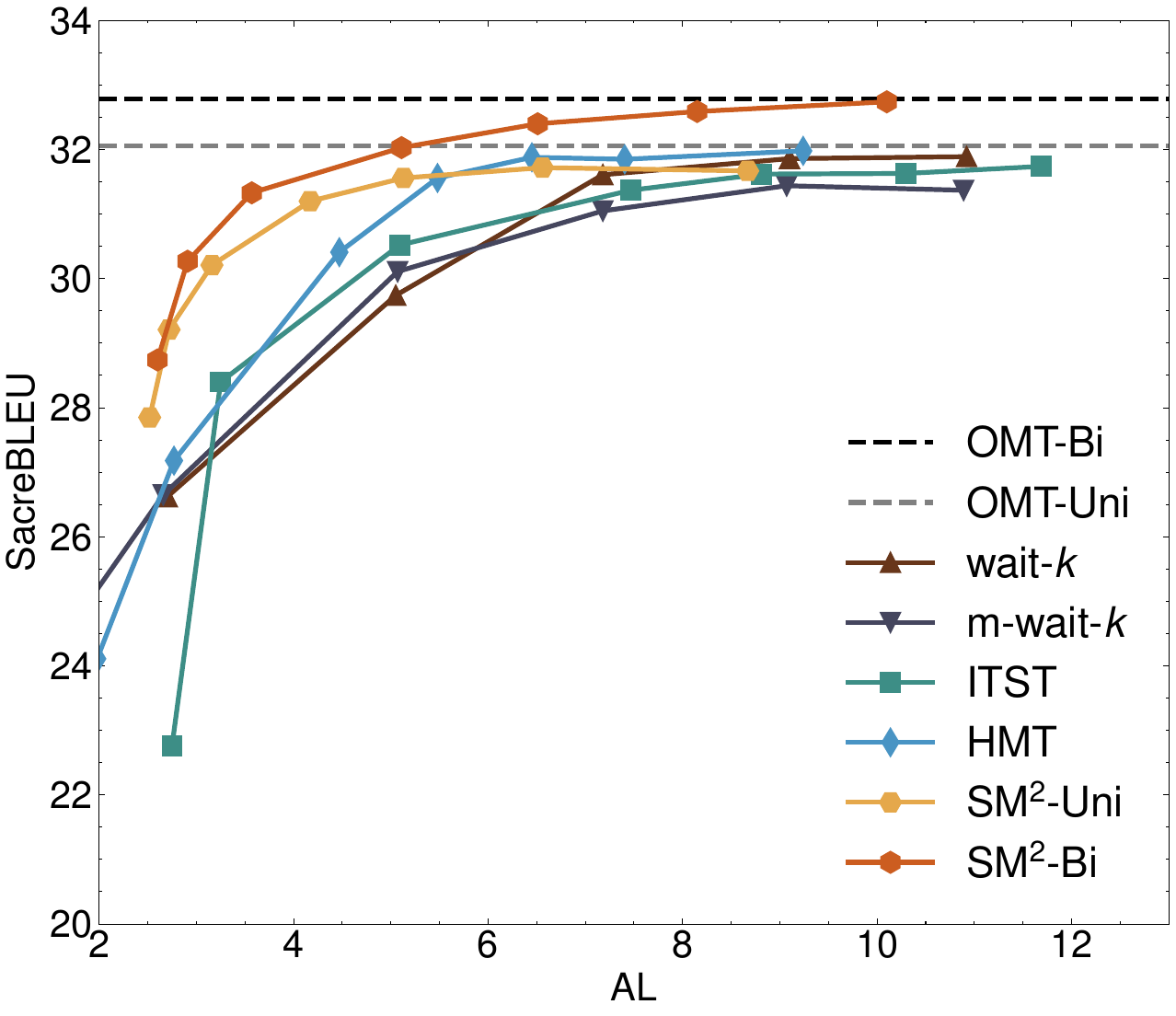} 
        }
	\caption{SacreBLEU against Average Lagging (AL) on Zh$\rightarrow$En, De$\rightarrow$En and En$\rightarrow$Ro.}
	\label{bleu-al}
    \end{figure*}
\begin{figure*}[h]
	\centering
	\small
        \subfigure[Zh$\rightarrow$En]{
            \includegraphics[width=0.3\textwidth]{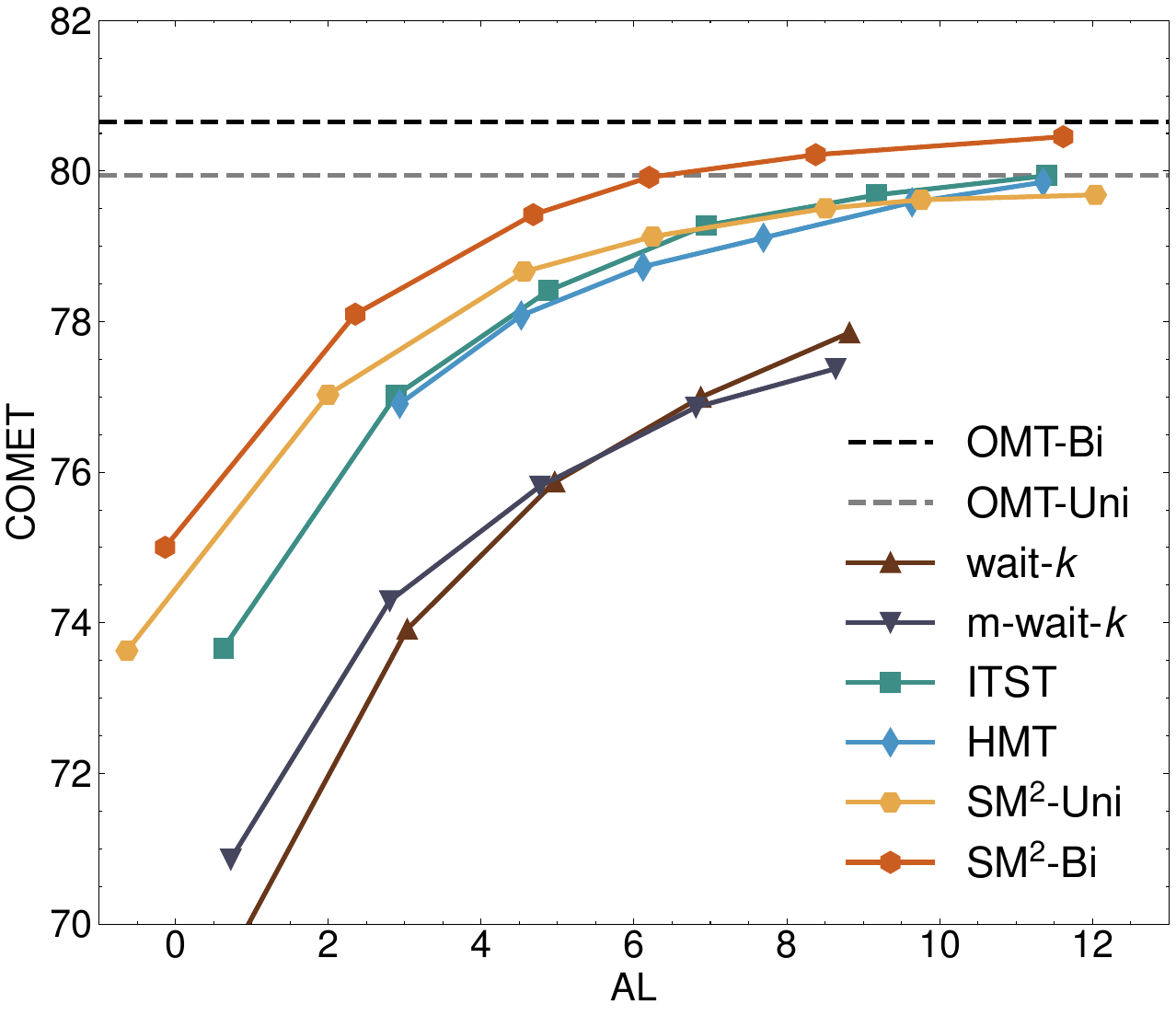} 
        }
        \subfigure[De$\rightarrow$En]{
            \includegraphics[width=0.3\textwidth]{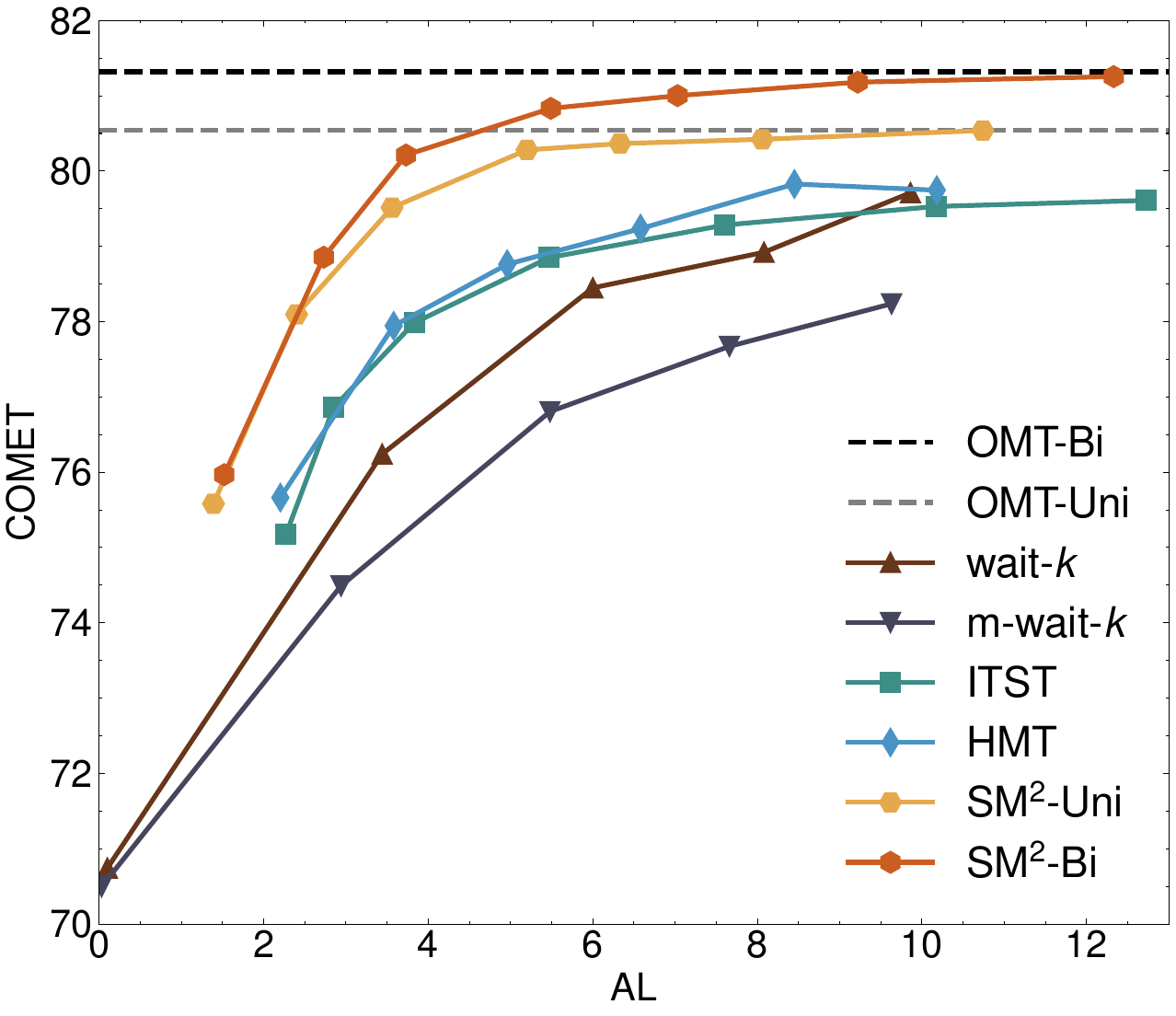}
        }
        \subfigure[En$\rightarrow$Ro]{
            \includegraphics[width=0.3\textwidth]{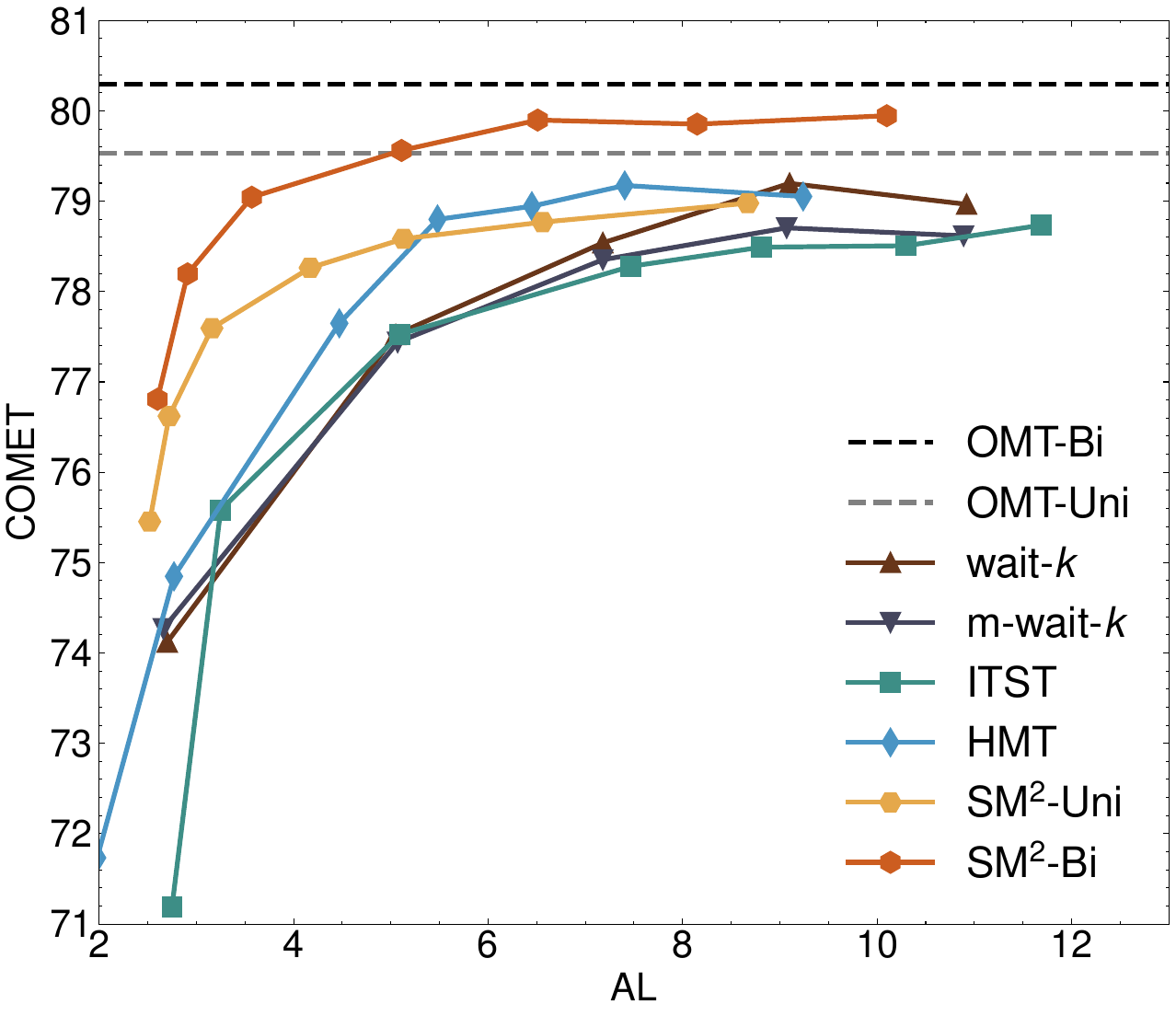} 
        }
	\caption{COMET against Average Lagging (AL)  on Zh$\rightarrow$En, De$\rightarrow$En and En$\rightarrow$Ro.}
	\label{comet-al}
    \end{figure*}
\section{Results and Analysis}

\subsection{Simultaneous Translation Quality}
We present the translation quality under various latency levels of different SiMT models in Figure \ref{bleu-al} and Figure \ref{comet-al}. These results indicate that SM$^2$ outperforms previous methods across three language pairs in terms of both SacreBLEU and COMET scores. With the unidirectional encoder, SM-Uni achieves higher translation quality compared to current state-of-the-art SiMT models (ITST, HMT) at low and medium latency levels (AL$ \in [0,6]$), and maintains comparable performance at high latency level (AL$ \in [6,12]$). We attribute this improvement to the effectiveness of learning a better policy during training. Furthermore, with the superior capabilities of the bidirectional encoder, SM$^2$-Bi outperforms previous SiMT models more significantly across all latency levels. All SiMT models with unidirectional encoders can approach the translation quality of OMT-Uni at high latency levels, but only SM$^2$-Bi achieves similar performance to OMT-Bi as the latency increases. These experimental results prove that SM$^2$ achieves better performance than other SiMT methods for learning better policy and improving translation quality. Detailed numerical results are provided in Appendix \ref{num-results}, supplemented with additional evidence demonstrating the robustness of SM$^2$ to sentence length variations.

\subsection{Superiority of SM$^2$ in Learning Policy}
\begin{figure}[t]
	\centering
	\small
        \includegraphics[width=0.47\textwidth]{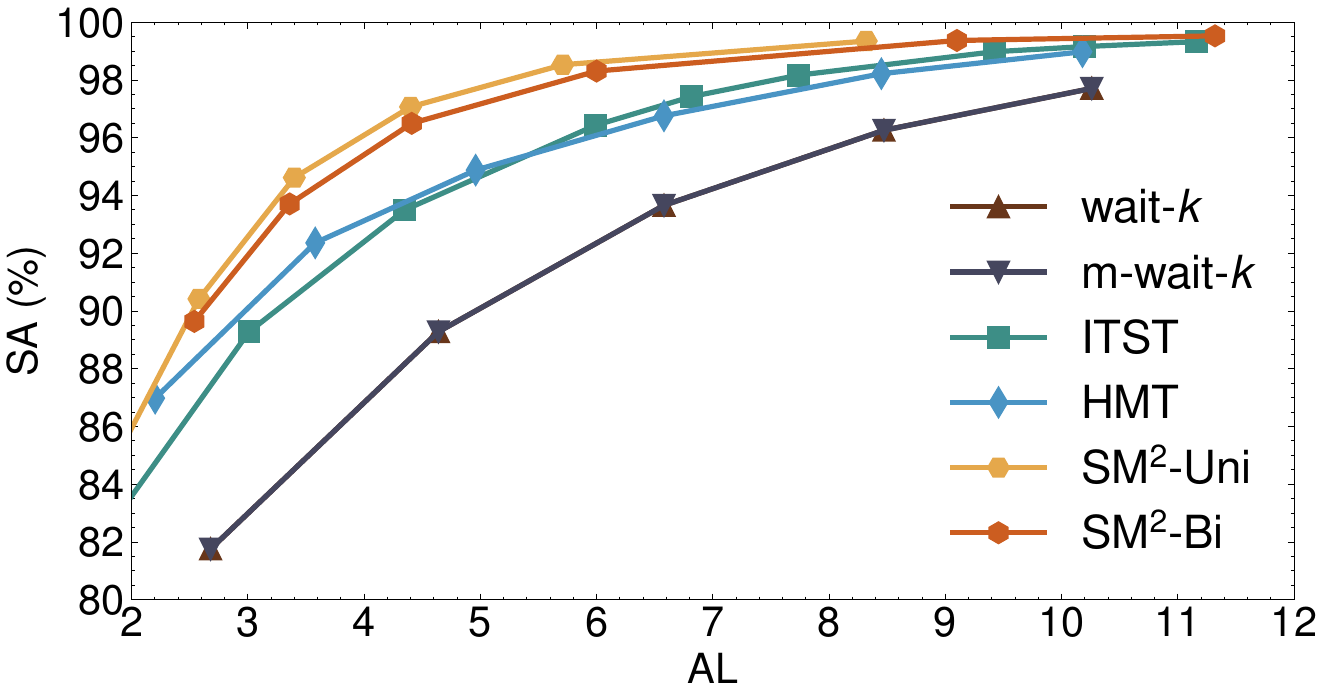}
	\caption{Evaluation of different SiMT policies. We calculate SA ($\uparrow$) under different latency levels.}
	\label{policy-eval}
    \end{figure}

To verify whether SM$^2$ can learn a more effective policy, we compare SA($\uparrow$) under various latency levels of different SiMT models. Following \citet{zhang-feng-2022-information} and \citet{kim2023enhanced}, we conduct the analysis on RWTH\footnote{\url{https://www-i6.informatik.rwth-aachen.de/goldAlignment/}}, a De$\rightarrow$En alignment dataset. The results are presented in Figure \ref{policy-eval}. Compared with existing methods, both SM$^2$-Uni and SM$^2$-Bi receive more aligned source tokens before generating target tokens under the same latency. Especially at medium latency level (AL$ \in [4,6]$), SM$^2$ can receive about 8\% more source tokens than fixed policies (wait-$k$, m-wait-$k$) and 3.6\% more than adaptive policies (ITST, HMT). We attribute these improvements to the advantages of SM$^2$ in learning policy. Through precise optimization, SM$^2$ can make more suitable decisions at each state, which generates faithful translations once receiving sufficient source tokens and waits for more source inputs when the predicted tokens are incredible. With sufficient exploration, SM$^2$ can investigate all possible situations and reduce unnecessary latency in the SiMT process.

\subsection{Precise Optimization for Each Decision}
\label{precise-optimization}
To validate whether the confidence-based policy is precisely optimized at each state, we examine the relationship between estimated confidence $c_{ij}$ and the probability of the correct token $y_{i}$ in the prediction, denoted as $p^c_{ij}$. Specifically, we employ SM$^2$ to decode the validation set in a teacher-forcing manner, calculating the $c_{ij}$ and $p^c_{ij}$ for all possible states. Subsequently, a correlation analysis is performed between $c_{ij}$ and $p^c_{ij}$. The results in Table \ref{conf-correlation} demonstrate a strong correlation, evidenced by high values in Pearson (0.82) and Spearman (0.84) coefficients, with a slightly moderate but significant Kendall's $\tau$ coefficient (0.65). These results suggest a robust linear and monotonic relationship between $c_{ij}$ and $p^c_{ij}$, indicating the capacity of $c_{ij}$ to accurately assess the credibility of the current predicted token. Consequently, this confirms the effectiveness of the confidence-based policy in making precise decisions at each state.

    \begin{table}[t]
            \setlength{\tabcolsep}{1mm}
            \setlength{\abovecaptionskip}{0.3cm}
            \setlength{\belowcaptionskip}{-0.3cm}
    	\centering
    	\small
    \resizebox{\columnwidth}{!}{%
    \begin{tabular}{c|ccc}
    \toprule
    \textbf{Correlation Coefficient}& Pearson & Spearman & Kendall's $\tau$ \\
    \midrule
    \textbf{Value} & 0.82 & 0.84 & 0.65 \\
    \bottomrule
    \end{tabular}%
    }
    \caption{Correlation between $c_{ij}$ and $p^c_{ij}$.}
    \label{conf-correlation}
    \end{table}

     \begin{figure}[t]
	\centering
	\small
        \subfigure[HMT]{
            \includegraphics[width=0.2\textwidth]{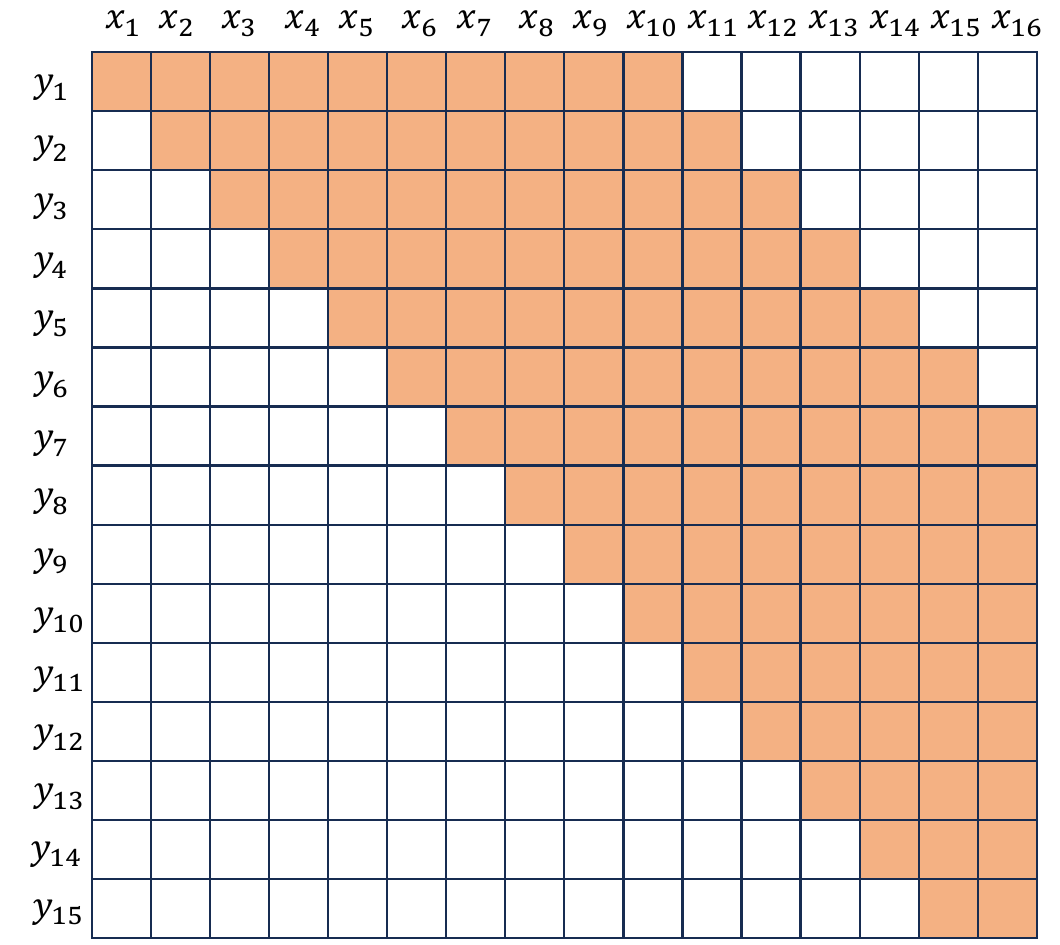} 
            \label{pre-prune}
        }
        \subfigure[SM$^2$]{
            \includegraphics[width=0.2\textwidth]{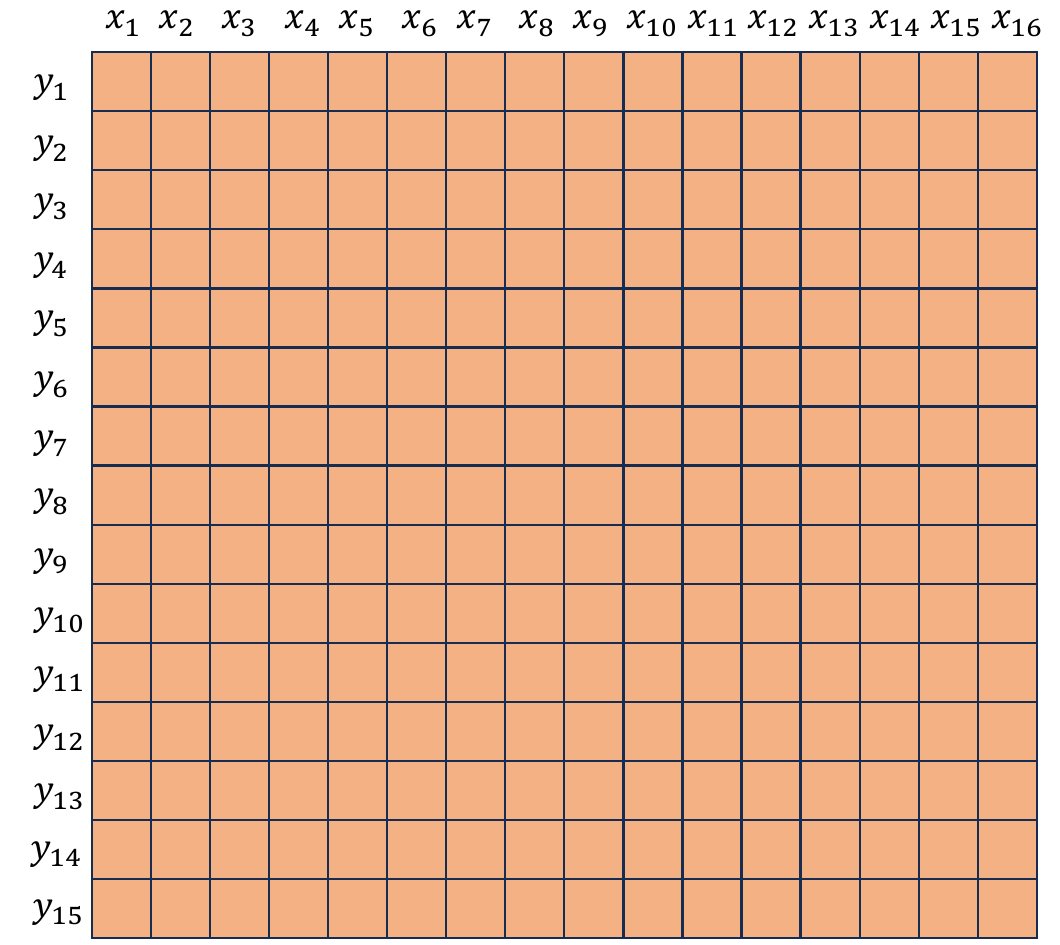}
            \label{all-space}
        }
        \subfigure[Effect of prohibition to the SiMT performance.]{
        \includegraphics[width=0.44\textwidth]{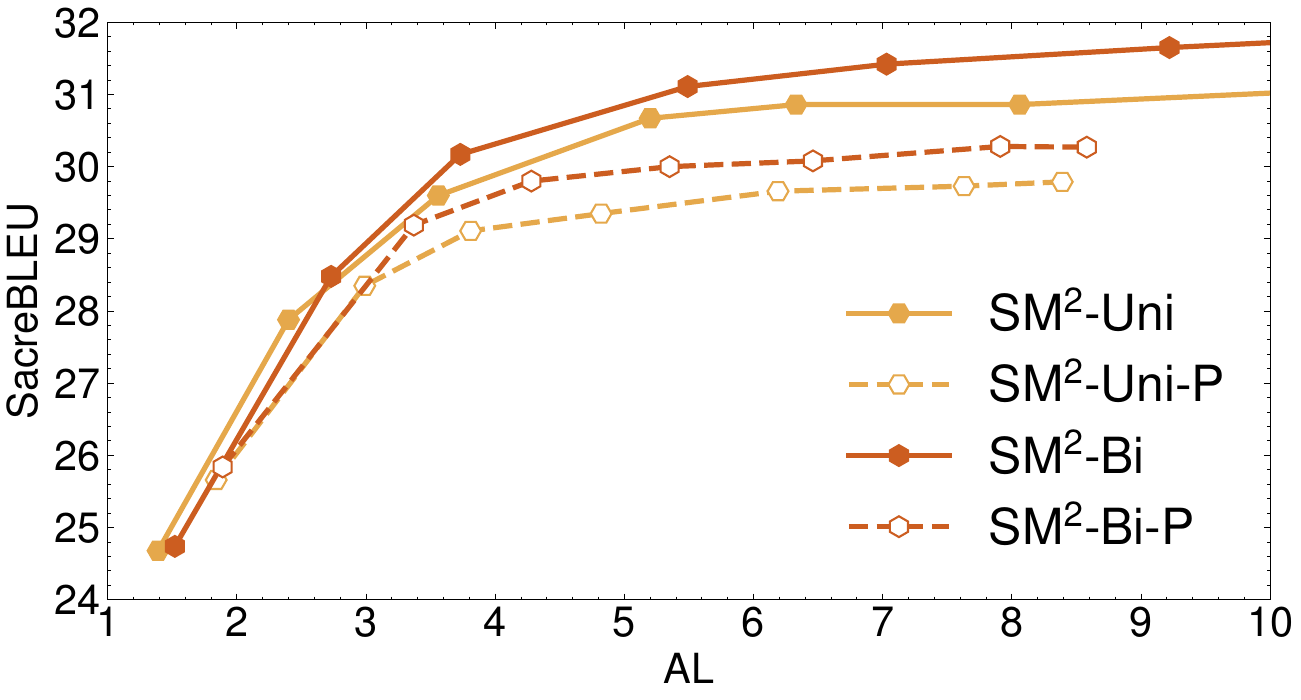}
        \label{policy-consts-bleu-al}
        }
	\caption{The visualization and effect of prohibition. In (a) and (b), the shaded areas represent the states allowed for exploration in training. We apply the same prohibition in HMT \cite{zhang2023hidden} to train SM$^2$-Uni-P and SM$^2$-Bi-P.}
	\label{search-space}
    \end{figure}
\subsection{Advantage of Sufficient Exploration} 
\label{sufficient-exploration}
Existing methods often prohibit the exploration of some paths due to the possible decision paths being numerous \citep{zheng2019simultaneous,miao2021generative,zhang2023hidden}. To investigate the impact of the prohibition on SiMT models and the superiority of SM$^2$ in sufficiently exploring all states, we attempt to train these methods without prohibition, but they fail to converge. Therefore, we analyze the impact by employing the same prohibition in HMT \cite{zhang2023hidden} and RIL\cite{zheng2019simultaneous} to train SM$^2$, which restricts SM$^2$ to explore states only between wait-$k_1$ and wait-$k_2$ paths in training. As shown in Figure \ref{pre-prune}, we set $k_1=1$ and $k_2=10$ in our experiments. The performances of SM$^2$ with prohibition (SM$^2$-Uni-P and SM$^2$-Bi-P) are shown in Figure \ref{policy-consts-bleu-al}, indicating a decline in performance. These results suggest that the prohibition causes insufficient exploration, leading to diminished performance. In contrast, SM$^2$ ensures comprehensive exploration, which is shown in Figure \ref{all-space}, thereby achieving higher performance. Further analysis of the policy quality is provided in Appendix \ref{restrict-sa}.

\subsection{Compatibility with OMT Models}
\begin{figure}[t]
	\centering
	\small
        \subfigure[ Comparison with SiMT models.]{
            \includegraphics[width=0.22\textwidth]{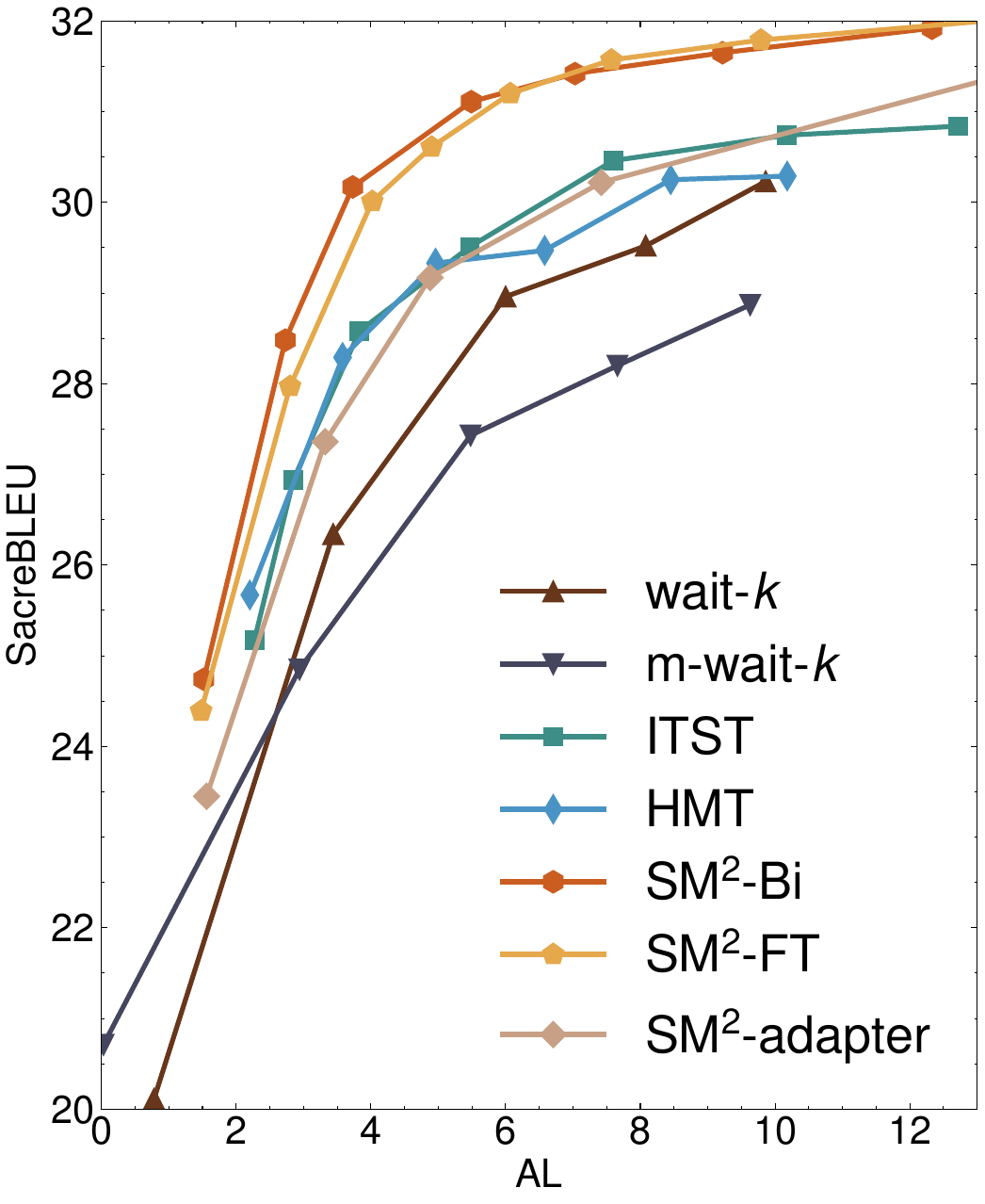} 
            \label{de-en-expand}
        }
        \subfigure[Comparison between different OMT models.]{
            \includegraphics[width=0.22\textwidth]{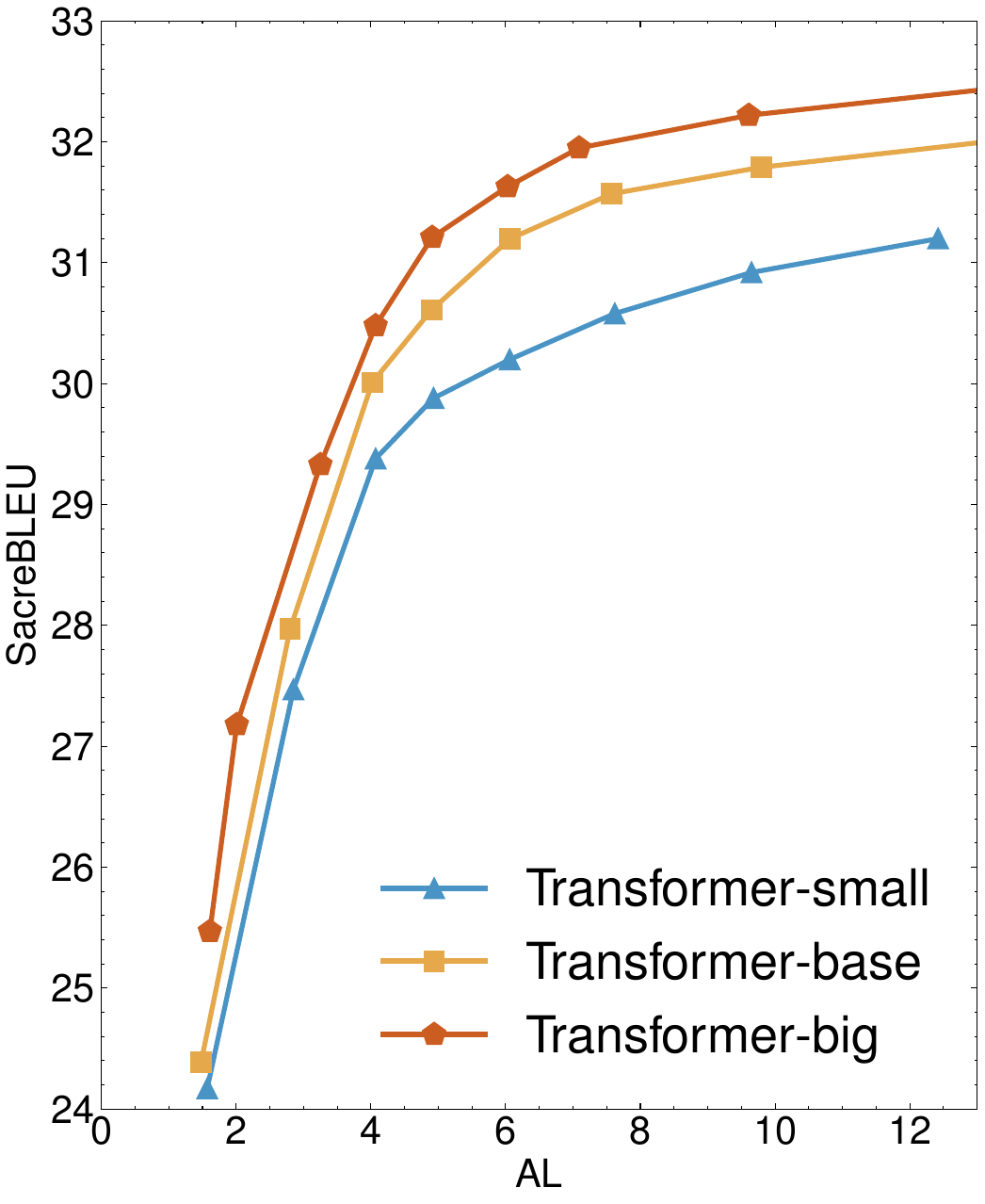}
            \label{de-en-size}
        }
	\caption{The SiMT performance of different OMT models after fine-tuning according to SM$^2$.}
	\label{de-en-omt-analysis}
    \end{figure}

SM$^2$ allows for the parallel training of the bidirectional encoder. Due to this compatibility, SM$^2$-Bi achieves superior translation quality than existing SiMT methods with unidirectional encoders (Figure \ref{bleu-al},\ref{comet-al}). To further present the superiority of this compatibility, we propose fine-tuning OMT models according to SM$^2$, so that the translation ability in OMT models can be easily utilized to gain SiMT models. Specifically, two distinct methods are used: fine-tuning all model parameters (SM$^2$-FT) and fine-tuning with adapters (SM$^2$-adapter)\footnote{We add adapters after the feed-forward networks of each encoder and decoder layer. For each adapter, the input dimension and output dimension are 512, and the hidden layer dimension is 128.}. As shown in Figure \ref{de-en-expand}, SM$^2$-adapter can achieve comparable performance with current state-of-the-art SiMT models, and SM$^2$-FT closely matches the performance of SM$^2$-Bi.

Additionally, we further explore the effect of the OMT models' translation abilities on the corresponding SiMT abilities after fine-tuning. We conduct the full-parameter fine-tuning on OMT models with Transformer-small, Transformer-base, and Transformer-big respectively. The OMT and SiMT capabilities of these models are illustrated in Table \ref{omt-ft-table} and Figure \ref{de-en-size}, which reveal that models with stronger OMT abilities achieve better SiMT performance after fine-tuning. Besides, the results in Table \ref{omt-ft-table} show that these models' original OMT abilities are not hurt, indicating that SM$^2$ enables models to support both OMT and SiMT abilities. 

\subsection{Ablation Study}
We conduct ablation studies on SM$^2$ to analyze the effect of $\mathcal{L}_{omt}$ and modification from OMT setting.

    \begin{table}
        \setlength{\tabcolsep}{1mm}
        \setlength{\abovecaptionskip}{0.3cm}
        \setlength{\belowcaptionskip}{-0.3cm}
	\centering
	\small
	\begin{tabular}{cccc}
\toprule
             & \multicolumn{1}{c}{}  & \multicolumn{2}{c}{\textbf{SacreBLEU}}  \\
        \cmidrule{3-4}
       \multirow{-2}{*}{\textbf{OMT model}} & \multicolumn{1}{c}{\multirow{-2}{*}{\textbf{Parameters}}}    & \textbf{before FT}    & \textbf{after FT}   \\ \midrule
Transformer-small& 47.9M& 30.86	& 31.33 \\
Transformer-base& 60.5M&31.93  & 31.87\\
Transformer-big& 209.1M&32.99 &	32.75\\ \bottomrule
	\end{tabular}
        \caption{The OMT performance of different OMT models before/after fine-tuning according to SM$^2$.}
        \label{omt-ft-table}
    \end{table}

\begin{figure}[t]
	\centering
	\small
        \includegraphics[width=0.45\textwidth]{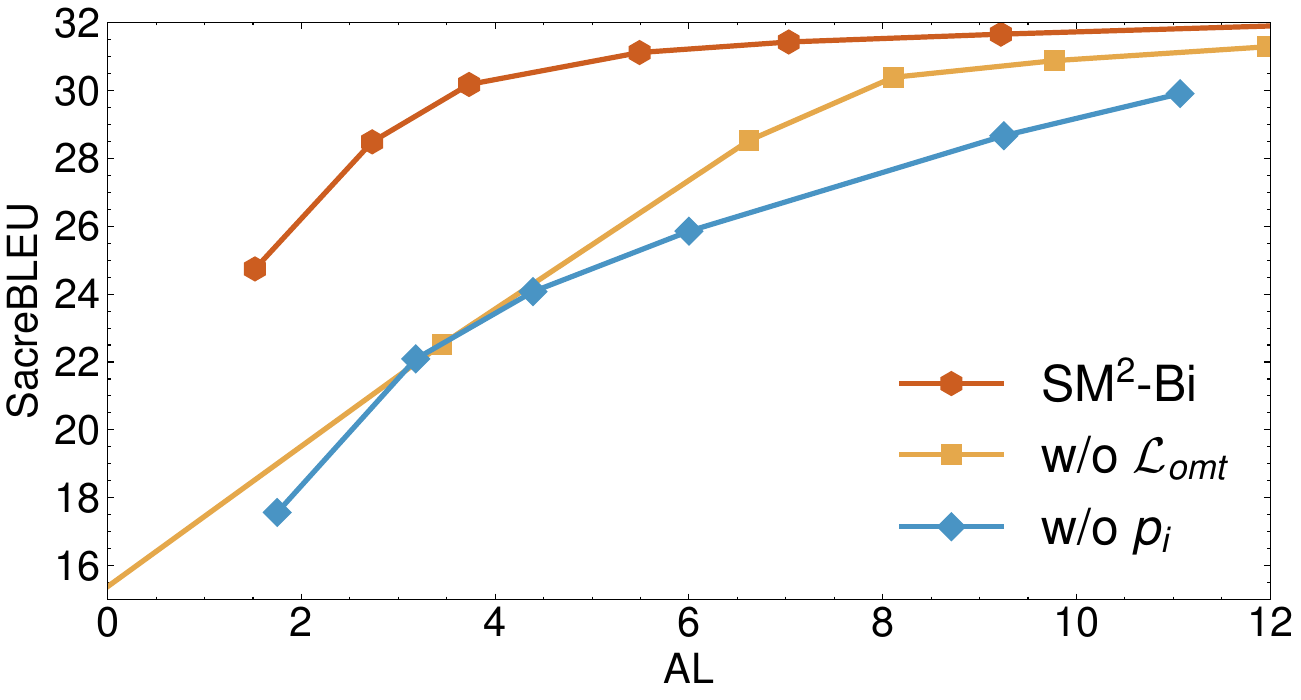}
	\caption{Effect of $\mathcal{L}_{omt}$ and modification from OMT setting on the SM$^2$."w/o $\mathcal{L}_{omt}$" is SM$^2$ trained without $\mathcal{L}_{omt}$, and "w/o $p_i$" means SM$^2$ trained using one-hot rather than OMT setting for modification.}
	\label{ablation-study}
    \end{figure}

\textbf{Effect of $\mathcal{L}_{omt}$} As shown in Figure \ref{ablation-study}, the SiMT model without $\mathcal{L}_{omt}$ drops quickly. We argue this is because training without $\mathcal{L}_{omt}$ may cause a worse modification. The results in Table \ref{ablation-study-omt} show that the OMT performance of SM$^2$ trained without $\mathcal{L}_{omt}$ is significantly affected, even worse than its SiMT performance in the high latency levels. This poor OMT ability cannot provide accurate modification, thus disrupting the policy learning process.

\textbf{Effect of OMT modification}
Following \textit{Ask For Hints} \citep{devries2018learning, lu2022learning}, we use the one-hot label as the "hints" to modify the prediction in SiMT setting. Specifically, we denote $t_i$ as the ground-truth label of the $i$-th target token, and hence the modification in SM$^2$ is adjusted as:
    \begin{equation}
        \begin{split}
        p^m_{ij}=c_{ij}\cdot p_{ij}+(1-c_{ij})\cdot t_i
        \end{split}
        \label{one-hot-modified}
    \end{equation}
As shown in Figure \ref{ablation-study}, the performance of SM$^2$ trained with modification in Eq.(\ref{one-hot-modified}) also drops. We argue that the modification from $t_i$ cannot reflect the real available gain from the modification after receiving the complete source sentence, thus learning a worse policy.

\label{sec:bibtex}
    \begin{table}[t]
        \setlength{\tabcolsep}{1mm}
        \setlength{\abovecaptionskip}{0.3cm}
        \setlength{\belowcaptionskip}{-0.3cm}
	\centering
	\small
	\begin{tabular}{c|ccc|c}
\toprule
     \textbf{Model}     & SM$^2$-Bi & w/o $\mathcal{L}_{omt}$ & w/o $p_{i}$ & OMT \\ \midrule
\textbf{SacreBLEU} & 31.87  & 30.33 & 31.95  & 31.93 \\\bottomrule
	\end{tabular}
        \caption{Effect of  $\mathcal{L}_{omt}$ and $p_{i}$ on the OMT ability.}
        \label{ablation-study-omt}
    \end{table}
\section{Related Work}
\noindent\textbf{Simultaneous Machine Translation} Different from offline machine translation \cite{vaswani2017attention,zhao2020knowledge,wu2024f}, existing SiMT methods are divided into fixed policy and adaptive policy. For fixed policy, \citet{ma2019stacl} proposed wait-$k$, which starts translation after receiving $k$ tokens. \citet{elbayad2020efficient} proposed multipath wait-$k$, which randomly samples $k$ during training. For adaptive policy, heuristic rules \citet{cho2016can} and reinforcement learning \citet{gu2017learning} are used to realize the SiMT task. \citet{mamonotonic} integrated multi-head monotonic attention to model the decision process, where each head independently makes decisions. Similarly, \citet{zhang2022gaussian} utilized Gaussian multi-head attention to model the alignment, thus improving the decision-making ability of each head. \citet{miao2021generative} proposed a generative framework to learn a read/write policy. \citet{zhang-feng-2022-information} measured the information SiMT had received and proposed an information-based policy. \citet{zhang2023hidden} used the Hidden Markov model in SiMT task to learn an adaptive policy.

Previous methods based on decision paths are limited in policy learning and model structure. Our proposed SM$^2$ individually explores all states during training, overcoming these limitations. 

\noindent\textbf{Confidence Estimation for OMT} Confidence estimation is used to measure the models' credibility. \citet{wang2019improving} used Monte Carlo dropout to propose an uncertainty-based confidence estimation. \citet{wan2020self} utilized the confidence score to guide self-paced learning. \citet{devries2018learning} evaluated the confidence by measuring the level it asks for hints from the ground-truth label, and \citet{lu2022learning} transferred it to OMT to improve the out-of-distribution detection.
\section{Conclusion}

In this paper, we propose \textbf{S}elf-\textbf{M}odifying \textbf{S}tate \textbf{M}odeling (SM$^2$), a novel training paradigm for SiMT. SM$^2$ eschews the construction of complete decision paths during training, opting to explore all potential states individually instead. By introducing the Self-Modifying process, SM$^2$ independently assesses each state to precisely optimize the read/write policy without the credit assignment problem. Through Prefix Sampling, SM$^2$ ensures sufficient exploration of all potential states. Experimental results across three language pairs validate the superior performance of SM$^2$, and our analyses further confirm that SM$^2$ can learn a more effective read/write policy. More promisingly, SM$^2$ demonstrates the potential to endow OMT models with SiMT capability through fine-tuning.
\section*{Limitations}
In this paper, we propose SM$^2$, a novel paradigm that individually optimizes decisions at each state. Although our experiments show the superiority of not building decision paths during training, there are still some parts to be further improved. For example, using a more effective way to independently assess the individual effect of each decision on the SiMT performance. Besides, how to leverage other pre-trained encoder-decoder models like BART and T5, to gain SiMT models, is still a promising direction to explore. These will be considered as objectives for our future work.

\section*{Acknowledgements}
This work has been supported by the National Natural Science Foundation of China (NSFC) under Grant No. 62206295 and 62206286. 
\bibliography{custom}

\begin{thebibliography}{33}
\expandafter\ifx\csname natexlab\endcsname\relax\def\natexlab#1{#1}\fi

\bibitem[{Cho and Esipova(2016)}]{cho2016can}
Kyunghyun Cho and Masha Esipova. 2016.
\newblock Can neural machine translation do simultaneous translation?
\newblock \emph{arXiv e-prints}, pages arXiv--1606.

\bibitem[{DeVries and Taylor(2018)}]{devries2018learning}
Terrance DeVries and Graham~W Taylor. 2018.
\newblock Learning confidence for out-of-distribution detection in neural networks.
\newblock \emph{arXiv preprint arXiv:1802.04865}.

\bibitem[{Elbayad et~al.(2020)Elbayad, Besacier, and Verbeek}]{elbayad2020efficient}
Maha Elbayad, Laurent Besacier, and Jakob Verbeek. 2020.
\newblock Efficient wait-k models for simultaneous machine translation.

\bibitem[{Grissom~II et~al.(2014)Grissom~II, He, Boyd-Graber, Morgan, and Daum{\'e}~III}]{grissom2014don}
Alvin Grissom~II, He~He, Jordan Boyd-Graber, John Morgan, and Hal Daum{\'e}~III. 2014.
\newblock Don’t until the final verb wait: Reinforcement learning for simultaneous machine translation.
\newblock In \emph{Proceedings of the 2014 Conference on empirical methods in natural language processing (EMNLP)}, pages 1342--1352.

\bibitem[{Gu et~al.(2017)Gu, Neubig, Cho, and Li}]{gu2017learning}
Jiatao Gu, Graham Neubig, Kyunghyun Cho, and Victor~OK Li. 2017.
\newblock Learning to translate in real-time with neural machine translation.
\newblock In \emph{Proceedings of the 15th Conference of the European Chapter of the Association for Computational Linguistics: Volume 1, Long Papers}, pages 1053--1062.

\bibitem[{Iranzo-S{\'a}nchez et~al.(2022)Iranzo-S{\'a}nchez, Civera, and Juan}]{iranzo2022simultaneous}
Javier Iranzo-S{\'a}nchez, Jorge Civera, and Alfons Juan. 2022.
\newblock From simultaneous to streaming machine translation by leveraging streaming history.
\newblock In \emph{Proceedings of the 60th Annual Meeting of the Association for Computational Linguistics (Volume 1: Long Papers)}, pages 6972--6985.

\bibitem[{Kang et~al.(2020)Kang, Zhao, Zhang, and Zong}]{kang2020dynamic}
Xiaomian Kang, Yang Zhao, Jiajun Zhang, and Chengqing Zong. 2020.
\newblock Dynamic context selection for document-level neural machine translation via reinforcement learning.
\newblock In \emph{Proceedings of the 2020 Conference on Empirical Methods in Natural Language Processing (EMNLP)}, pages 2242--2254.

\bibitem[{Kim and Cho(2023)}]{kim2023enhanced}
Kang Kim and Hankyu Cho. 2023.
\newblock Enhanced simultaneous machine translation with word-level policies.
\newblock \emph{arXiv preprint arXiv:2310.16417}.

\bibitem[{Lu et~al.(2022)Lu, Zeng, Zhang, Wu, and Li}]{lu2022learning}
Yu~Lu, Jiali Zeng, Jiajun Zhang, Shuangzhi Wu, and Mu~Li. 2022.
\newblock Learning confidence for transformer-based neural machine translation.
\newblock In \emph{Proceedings of the 60th Annual Meeting of the Association for Computational Linguistics (Volume 1: Long Papers)}, pages 2353--2364.

\bibitem[{Ma et~al.(2019)Ma, Huang, Xiong, Zheng, Liu, Zheng, Zhang, He, Liu, Li et~al.}]{ma2019stacl}
Mingbo Ma, Liang Huang, Hao Xiong, Renjie Zheng, Kaibo Liu, Baigong Zheng, Chuanqiang Zhang, Zhongjun He, Hairong Liu, Xing Li, et~al. 2019.
\newblock Stacl: Simultaneous translation with implicit anticipation and controllable latency using prefix-to-prefix framework.
\newblock In \emph{Proceedings of the 57th Annual Meeting of the Association for Computational Linguistics}, pages 3025--3036.

\bibitem[{Ma et~al.(2020{\natexlab{a}})Ma, Zhang, and Zhou}]{ma2020simple}
Shuming Ma, Dongdong Zhang, and Ming Zhou. 2020{\natexlab{a}}.
\newblock A simple and effective unified encoder for document-level machine translation.
\newblock In \emph{Proceedings of the 58th annual meeting of the association for computational linguistics}, pages 3505--3511.

\bibitem[{Ma et~al.(2020{\natexlab{b}})Ma, Pino, Cross, Puzon, and Gu}]{mamonotonic}
Xutai Ma, Juan~Miguel Pino, James Cross, Liezl Puzon, and Jiatao Gu. 2020{\natexlab{b}}.
\newblock Monotonic multihead attention.
\newblock In \emph{International Conference on Learning Representations}.

\bibitem[{Miao et~al.(2021)Miao, Blunsom, and Specia}]{miao2021generative}
Yishu Miao, Phil Blunsom, and Lucia Specia. 2021.
\newblock A generative framework for simultaneous machine translation.
\newblock In \emph{Proceedings of the 2021 Conference on Empirical Methods in Natural Language Processing}, pages 6697--6706.

\bibitem[{Minsky(1961)}]{minsky1961steps}
Marvin Minsky. 1961.
\newblock Steps toward artificial intelligence.
\newblock \emph{Proceedings of the IRE}, 49(1):8--30.

\bibitem[{Neishi and Yoshinaga(2019)}]{neishi2019relation}
Masato Neishi and Naoki Yoshinaga. 2019.
\newblock On the relation between position information and sentence length in neural machine translation.
\newblock In \emph{Proceedings of the 23rd Conference on Computational Natural Language Learning (CoNLL)}, pages 328--338.

\bibitem[{Post(2018)}]{post2018call}
Matt Post. 2018.
\newblock A call for clarity in reporting bleu scores.
\newblock In \emph{Proceedings of the Third Conference on Machine Translation: Research Papers}, page 186. Association for Computational Linguistics.

\bibitem[{Rei et~al.(2020)Rei, Stewart, Farinha, and Lavie}]{rei2020comet}
Ricardo Rei, Craig Stewart, Ana~C Farinha, and Alon Lavie. 2020.
\newblock Comet: A neural framework for mt evaluation.
\newblock In \emph{Proceedings of the 2020 Conference on Empirical Methods in Natural Language Processing (EMNLP)}, pages 2685--2702.

\bibitem[{Sennrich et~al.(2016)Sennrich, Haddow, and Birch}]{sennrich2016neural}
Rico Sennrich, Barry Haddow, and Alexandra Birch. 2016.
\newblock Neural machine translation of rare words with subword units.
\newblock In \emph{Proceedings of the 54th Annual Meeting of the Association for Computational Linguistics (Volume 1: Long Papers)}, pages 1715--1725.

\bibitem[{Vari{\v{s}} and Bojar(2021)}]{varivs2021sequence}
Du{\v{s}}an Vari{\v{s}} and Ond{\v{r}}ej Bojar. 2021.
\newblock Sequence length is a domain: Length-based overfitting in transformer models.
\newblock \emph{arXiv preprint arXiv:2109.07276}.

\bibitem[{Vaswani et~al.(2017)Vaswani, Shazeer, Parmar, Uszkoreit, Jones, Gomez, Kaiser, and Polosukhin}]{vaswani2017attention}
Ashish Vaswani, Noam Shazeer, Niki Parmar, Jakob Uszkoreit, Llion Jones, Aidan~N Gomez, {\L}ukasz Kaiser, and Illia Polosukhin. 2017.
\newblock Attention is all you need.
\newblock \emph{Advances in neural information processing systems}, 30.

\bibitem[{Wan et~al.(2020)Wan, Yang, Wong, Zhou, Chao, Zhang, and Chen}]{wan2020self}
Yu~Wan, Baosong Yang, Derek~F Wong, Yikai Zhou, Lidia~S Chao, Haibo Zhang, and Boxing Chen. 2020.
\newblock Self-paced learning for neural machine translation.
\newblock \emph{arXiv preprint arXiv:2010.04505}.

\bibitem[{Wang et~al.(2019)Wang, Liu, Wang, Luan, and Sun}]{wang2019improving}
Shuo Wang, Yang Liu, Chao Wang, Huanbo Luan, and Maosong Sun. 2019.
\newblock Improving back-translation with uncertainty-based confidence estimation.
\newblock \emph{arXiv preprint arXiv:1909.00157}.

\bibitem[{Wu et~al.(2024)Wu, Liu, and Zong}]{wu2024f}
Junhong Wu, Yuchen Liu, and Chengqing Zong. 2024.
\newblock F-malloc: Feed-forward memory allocation for continual learning in neural machine translation.
\newblock \emph{arXiv preprint arXiv:2404.04846}.

\bibitem[{Zhang et~al.(2020)Zhang, Zhang, He, Wu, and Wang}]{zhang2020learning}
Ruiqing Zhang, Chuanqiang Zhang, Zhongjun He, Hua Wu, and Haifeng Wang. 2020.
\newblock Learning adaptive segmentation policy for simultaneous translation.
\newblock In \emph{Proceedings of the 2020 Conference on Empirical Methods in Natural Language Processing (EMNLP)}, pages 2280--2289.

\bibitem[{Zhang and Feng(2021)}]{zhang2021universal}
Shaolei Zhang and Yang Feng. 2021.
\newblock Universal simultaneous machine translation with mixture-of-experts wait-k policy.
\newblock In \emph{Proceedings of the 2021 Conference on Empirical Methods in Natural Language Processing}, pages 7306--7317.

\bibitem[{Zhang and Feng(2022{\natexlab{a}})}]{zhang2022gaussian}
Shaolei Zhang and Yang Feng. 2022{\natexlab{a}}.
\newblock Gaussian multi-head attention for simultaneous machine translation.
\newblock \emph{arXiv preprint arXiv:2203.09072}.

\bibitem[{Zhang and Feng(2022{\natexlab{b}})}]{zhang-feng-2022-information}
Shaolei Zhang and Yang Feng. 2022{\natexlab{b}}.
\newblock \href {https://aclanthology.org/2022.emnlp-main.65} {Information-transport-based policy for simultaneous translation}.
\newblock In \emph{Proceedings of the 2022 Conference on Empirical Methods in Natural Language Processing}, pages 992--1013, Abu Dhabi, United Arab Emirates. Association for Computational Linguistics.

\bibitem[{Zhang and Feng(2022{\natexlab{c}})}]{zhang2022modeling}
Shaolei Zhang and Yang Feng. 2022{\natexlab{c}}.
\newblock Modeling dual read/write paths for simultaneous machine translation.
\newblock In \emph{Proceedings of the 60th Annual Meeting of the Association for Computational Linguistics (Volume 1: Long Papers)}, pages 2461--2477.

\bibitem[{Zhang and Feng(2023)}]{zhang2023hidden}
Shaolei Zhang and Yang Feng. 2023.
\newblock Hidden markov transformer for simultaneous machine translation.
\newblock \emph{arXiv preprint arXiv:2303.00257}.

\bibitem[{Zhang et~al.(2023)Zhang, Zhang, Liang, Xiang, Zhao, Zhou, and Zong}]{zhang2023layoutdit}
Zhiyang Zhang, Yaping Zhang, Yupu Liang, Lu~Xiang, Yang Zhao, Yu~Zhou, and Chengqing Zong. 2023.
\newblock Layoutdit: Layout-aware end-to-end document image translation with multi-step conductive decoder.
\newblock In \emph{Findings of the Association for Computational Linguistics: EMNLP 2023}, pages 10043--10053.

\bibitem[{Zhao et~al.(2023)Zhao, Fan, Luo, Jing, Wang, Zeng, and Huang}]{zhao2023adaptive}
Libo Zhao, Kai Fan, Wei Luo, Wu~Jing, Shushu Wang, Ziqian Zeng, and Zhongqiang Huang. 2023.
\newblock Adaptive policy with wait-k model for simultaneous translation.
\newblock In \emph{Proceedings of the 2023 Conference on Empirical Methods in Natural Language Processing}, pages 4816--4832.

\bibitem[{Zhao et~al.(2020)Zhao, Xiang, Zhu, Zhang, Zhou, and Zong}]{zhao2020knowledge}
Yang Zhao, Lu~Xiang, Junnan Zhu, Jiajun Zhang, Yu~Zhou, and Chengqing Zong. 2020.
\newblock Knowledge graph enhanced neural machine translation via multi-task learning on sub-entity granularity.
\newblock In \emph{Proceedings of the 28th International Conference on Computational Linguistics}, pages 4495--4505.

\bibitem[{Zheng et~al.(2019)Zheng, Zheng, Ma, and Huang}]{zheng2019simultaneous}
Baigong Zheng, Renjie Zheng, Mingbo Ma, and Liang Huang. 2019.
\newblock Simultaneous translation with flexible policy via restricted imitation learning.
\newblock \emph{arXiv preprint arXiv:1906.01135}.

\end{thebibliography}

\appendix
\section{Comparison between SM$^2$ and RL-based SiMT methods}
\label{comp-sm2-rl}
In the following, we will compare the similarities and differences between our proposed SM$^2$ and RL-based methods.

On the one hand, both SM$^2$ and RL-based methods train SiMT models to learn read/write actions at each state $s_{ij}$. Specifically, the read/write policy $\pi(a_{ij}\mid s_{ij})$ in SM$^2$ can be described as:
    \begin{equation}
        \begin{split}
        \pi(a_{ij}\mid s_{ij})=
        \begin{cases}
        c_{ij} & a_{ij}={\textrm{WRITE}} \\
        1-c_{ij} & a_{ij}={\textrm{READ}} 
        \end{cases}
        \end{split}
        \label{sm2-rl-policy}
    \end{equation}
For each state $s_{ij}$, the reward in SM$^2$ can be described as:
    \begin{equation}
        \begin{split}
           r_{ij}=y_i\log(p^m_{ij})
        \end{split}
        \label{sm2-rl-policy}
    \end{equation}
During training, the policy can be optimized based on the reward and converge to the optimal policy.

On the other hand, SM$^2$ offers additional advantages over RL-based methods. Firstly, the reward in SM$^2$ is differentiable, allowing the policy to be optimized by directly using the reward as the objective. In contrast, the reward in RL-based methods \citep{grissom2014don,gu2017learning} is undifferentiable, which can hinder stable training. Secondly, SM$^2$ independently assesses each state, avoiding the credit assignment problem in existing RL-based methods.

\section{Gradient Analysis}
\label{gradient-analysis}
In this section, we provide a gradient analysis of the independent optimization in SM$^2$. The training loss function $\mathcal{L}$ of SM$^2$ is formulated in Eq. (\ref{group-loss}). During training, this loss function adjusts each decision $d_{ij}$ at state $s_{ij}$ by changing the value of corresponding confidence $c_{ij}$.
Specifically, the gradient of $\mathcal{L}$ with respect to $c_{ij}$ is calculated as:
    \begin{equation}
        \begin{split}
         \frac{\partial\mathcal{L}}{\partial c_{ij}}&=\frac{\partial\mathcal{L}_{s_{ij}}}{\partial c_{ij}}+\lambda\frac{\partial\mathcal{L}_{c_{ij}}}{\partial c_{ij}} \\
         &=-\frac{y_{i}}{p^m_{ij}}\cdot\frac{\partial p^m_{ij}}{\partial c_{ij}}-\frac{\lambda}{c_{ij}}\\
         &=-\frac{y_{i}(p_{ij}-p_i)}{c_{ij}\cdot p_{ij}+(1-c_{ij})\cdot p_i}-\frac{\lambda}{c_{ij}}
        \end{split}
        \label{grad-analysis-sm2}
    \end{equation}
 It is evident that this gradient does not contain any 
 $c_{i'j'}$ ($i'\neq i$ or $j'\neq j$). Therefore, in the training process, the estimated value of $c_{ij}$ is adjusted only based on its current value and the prediction probability of the current state, without being affected by the decisions at other states, thus allowing for the independent optimization of $c_{ij}$.
 
 In contrast, existing SiMT methods usually conduct training on decision paths and can not ensure independent optimization. Taking ITST \cite{zhang-feng-2022-information} as an example, whose loss function $\mathcal{L}'$ is formulated as:
    \begin{equation}
        \begin{split}
\mathcal{L}'&=\mathcal{L}_{ce} + \mathcal{L}_{latency} +  \mathcal{L}_{norm}\\
\mathcal{L}_{latency}&=\sum_{i=1}^{I}\sum_{j=1}^{J}\;T_{ij}\times C_{ij}\\
\mathcal{L}_{norm}&=\sum_{i=1}^{I} \left\|\sum_{j=1}^{J}T_{ij}-1  \right\|_{2}
        \end{split}
        \label{loss-itst}
    \end{equation}
where $\mathcal{L}_{ce}$ is the cross-entropy for learning translation ability, and $C_{ij}$ is the latency cost for each state. During training, the decision is dominated by $T_{ij}$. The gradient of $\mathcal{L}'$ with respect to $T_{ij}$ is calculated as:
    \begin{equation}
        \begin{split}
\frac{\partial \mathcal{L}'}{\partial T_{ij}}&=\frac{\partial \mathcal{L}_{ce}}{\partial T_{ij}}+C_{ij}+2(\sum_{j=1}^{J}T_{ij}-1)\\
        \end{split}
        \label{grad-analysis-itst}
    \end{equation}
It is noted that the gradient of $T_{ij}$ is also affected by the current values of $T_{ij'}$($j'=1,2,...,J$). These decisions are coupled in the optimization, thus not enabling the independent optimization of each decision. This can trigger mutual interference during training \cite{zhang2023hidden} and lead to a credit assignment problem.
\section{Effect of $\lambda$}
\label{effect-of-lambda}
    \begin{figure}
	\centering
	\small
        \subfigure[Performance of SM$^2$ with different $\lambda$.]{
            \includegraphics[width=0.43\textwidth]{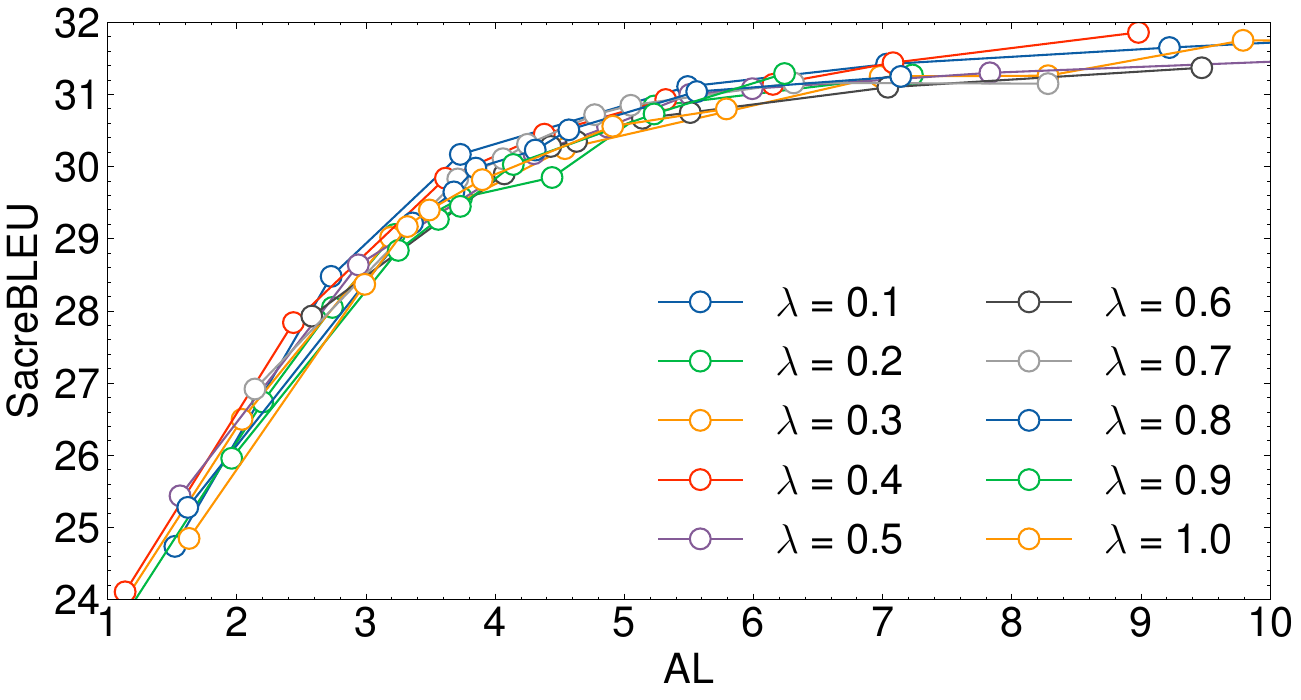} 
            \label{lambda-effect-bleu-al}
        }
        \subfigure[Max Latency of SM$^2$ with different $\lambda$.]{
        \includegraphics[width=0.43\textwidth]{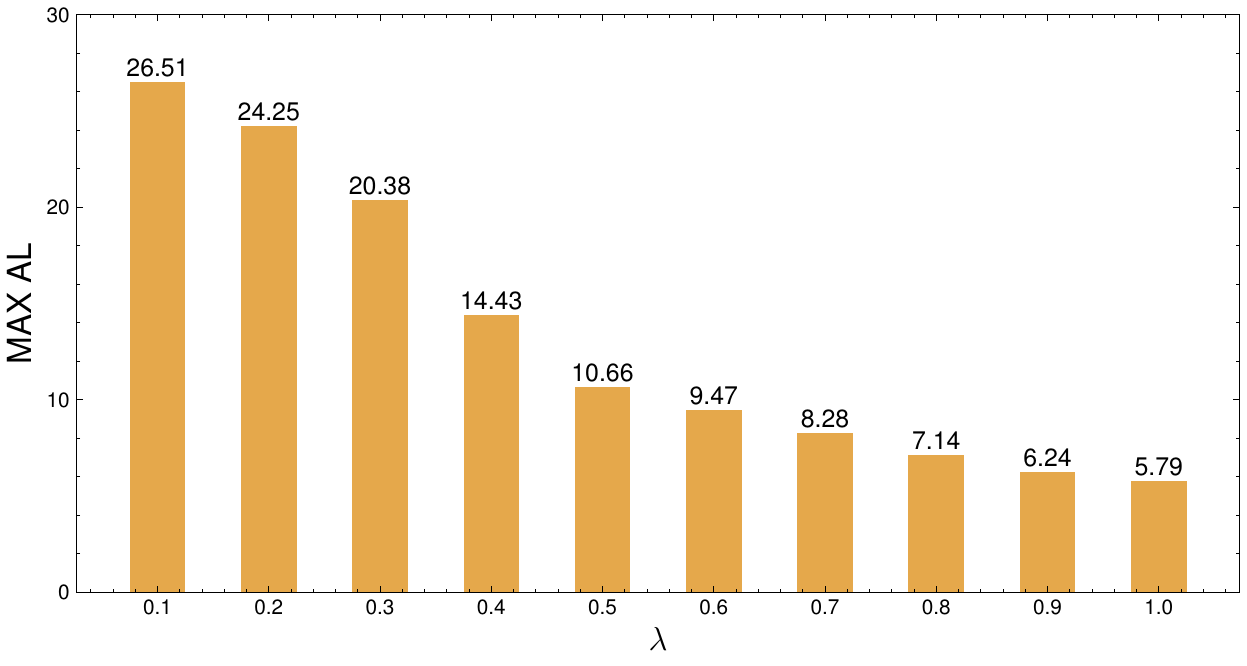}
        \label{lambda-effect-max-al}
        }
	\caption{Effect of $\lambda$ on SM$^2$.}
	\label{lambda-effect}
    \end{figure}
We analyze the effect of $\lambda$, which is the weight of the penalty during training. We train SM$^2$ with different $\lambda$ ranging from 0.1 to 1, in increments of 0.1. As shown in Figure \ref{lambda-effect-bleu-al}, the SM$^2$ models trained with different $\lambda$ show comparable performance across all latency. This indicates that SM$^2$ is robust to variations in hyper-parameters $\lambda$.

When $\lambda$ becomes larger, the corresponding $\gamma$ at the same latency will also increase. Therefore, we further analyze the effect of $\lambda$ on the applicable latency range of SM$^2$. We denote the "MAX AL" as the latency of SM$^2$ when $\gamma$ is set as 0.99 during inference. The results are shown in Figure \ref{lambda-effect-max-al}. When $\lambda$ becomes larger, "MAX AL" also decreases, which means a smaller applicable latency range. For example, when $\lambda=1.0$ in training, it is hard for SM$^2$ to perform SiMT task under the latency levels where AL is larger than 5.79 since the threshold $\gamma$ has been close to 1.
\section{Hyper-parameters}
\label{hyper}
The system settings in our experiments are shown in Table \ref{Hyper}. We set $\lambda=0.1$ during training. Besides, we follow \citet{mamonotonic} to use greedy search during inference for all baselines. The values of $\gamma$ we used are 0.3,0.4,0.5,0.55,0.6,0.65 for Zh$\rightarrow$En, 0.3,0.4,0.5,0.55,0.6,0.65,0.7 for De$\rightarrow$En, and 0.3,0.4,0.5,0.6,0.65,0.7,0.75 for En$\rightarrow$Ro.
\begin{table}[t]
\centering
\begin{tabular}{l c }
\toprule
\multicolumn{2}{c}{\textbf{Hyper-parameter}} \\
\midrule[1pt]
encoder layers          & 6                              \\
encoder attention heads & 8                             \\
encoder embed dim       & 512                      \\
encoder ffn embed dim   & 1024                     \\
decoder layers          & 6                        \\
decoder attention heads & 8                    \\
decoder embed dim       & 512                     \\
decoder ffn embed dim   & 1024                     \\
dropout                 & 0.1                 \\
optimizer               & adam  \\
adam-$\beta$          & (0.9, 0.98)    \\
clip-norm               & 1e-7                \\
lr                      & 5e-4                \\
lr scheduler        & inverse sqrt  \\
warmup-updates          & 4000          \\
warmup-init-lr          & 1e-7     \\
weight decay            & 0.0001        \\
label-smoothing         & 0.1                \\
max tokens              & 8192 \\
\bottomrule
\end{tabular}
\caption{Hyper-parameters of our experiments.}
\label{Hyper}
\end{table}
\section{Main Results Supplement}
\label{num-results}
\subsection{Numerical Results}
\begin{figure}[t]
    \centering
    \small
        \includegraphics[width=0.4\textwidth]{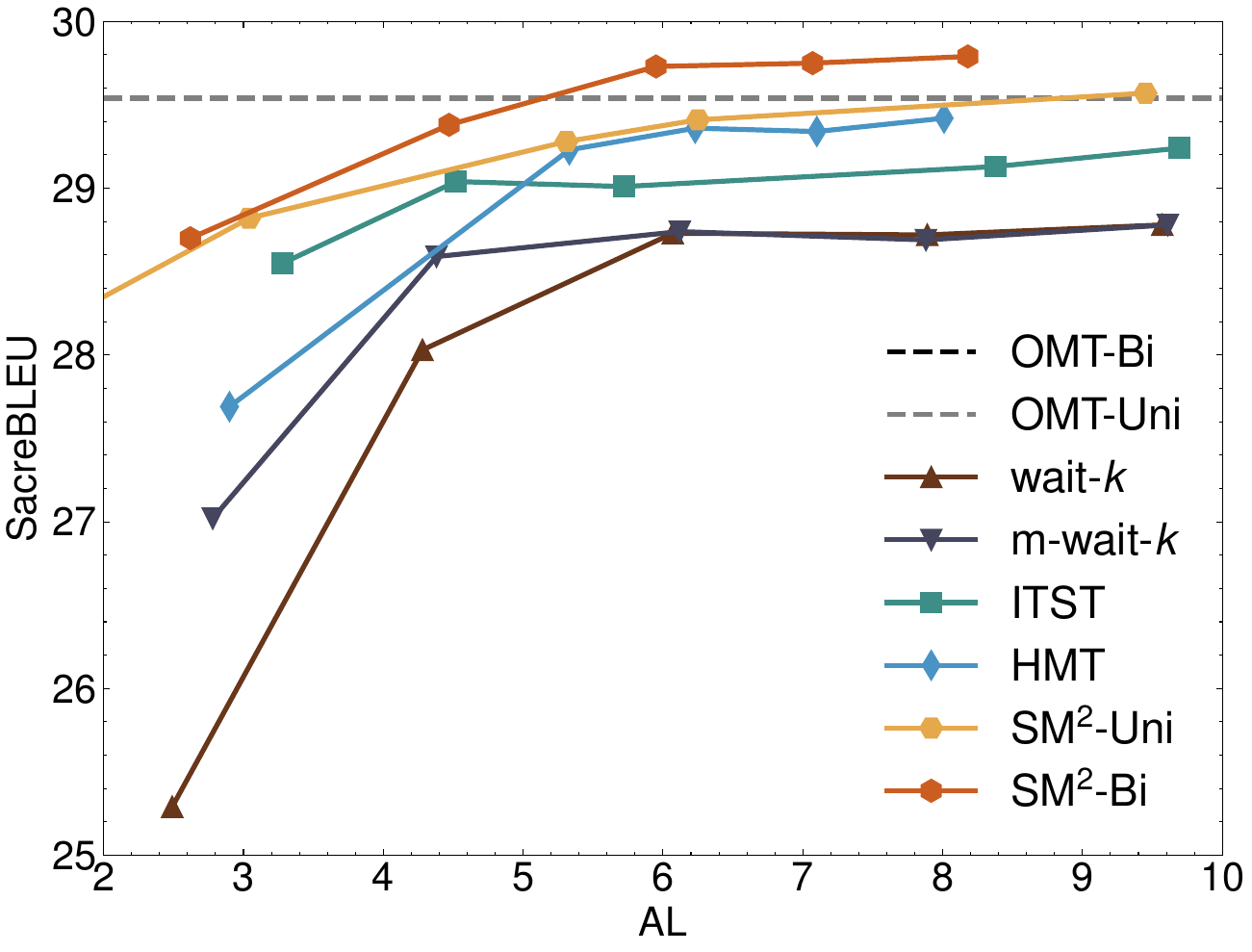}
    \caption{SacreBLEU against Average Lagging (AL) on En$\rightarrow$Vi}
    \label{en-vi-bleu-al}
    \end{figure}

\begin{figure}[t]
    \centering
    \small
        \includegraphics[width=0.4\textwidth]{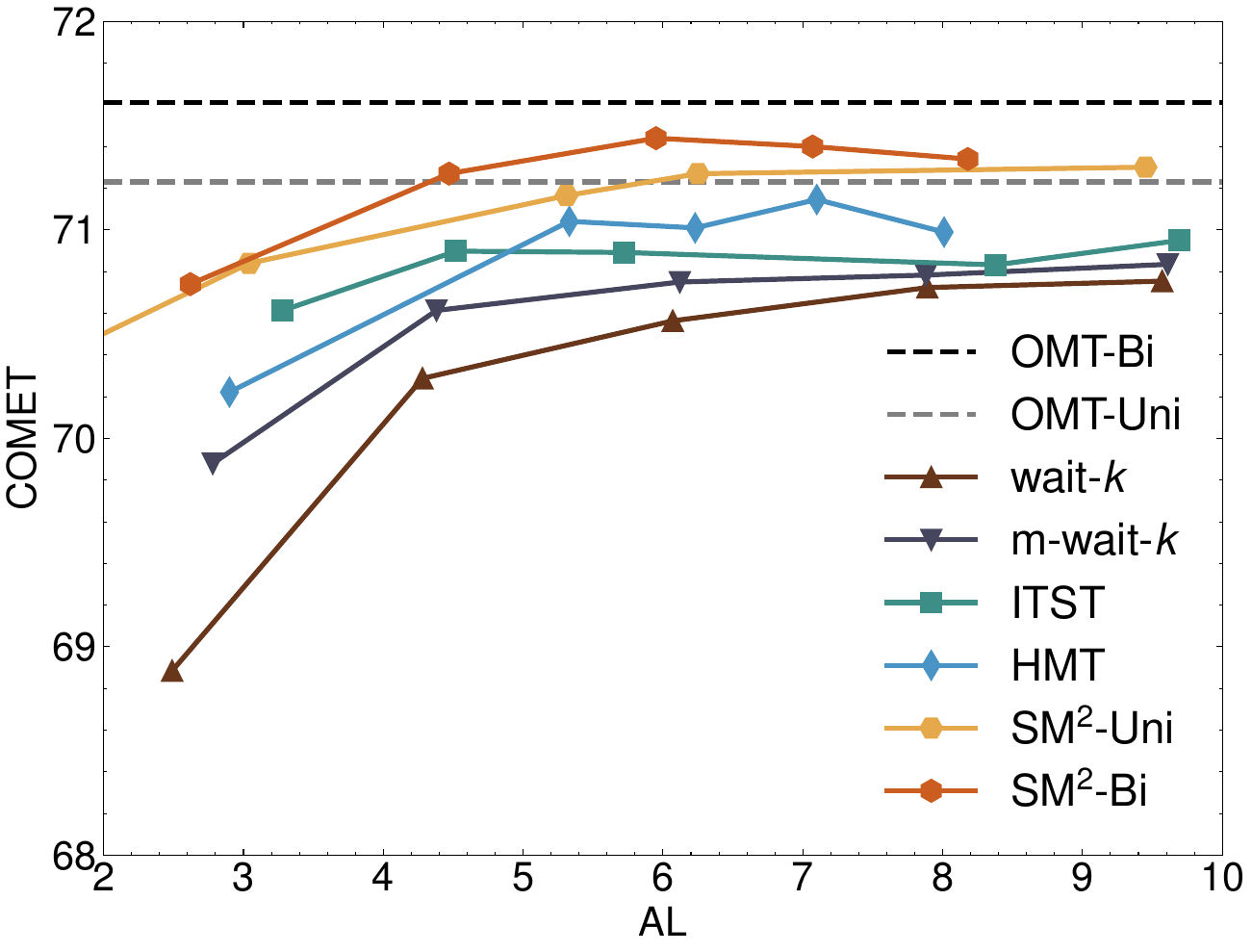}
    \caption{COMET against Average Lagging (AL) on En$\rightarrow$Vi}
    \label{en-vi-comet-al}
    \end{figure}

Table \ref{zh-en-num-results}, \ref{de-en-num-results}, \ref{en-ro-num-results} respectively report the numerical results on LDC Zh$\rightarrow $En, WMT15 De$\rightarrow $En, WMT16 En$\rightarrow $Ro measured by AL, SacreBLEU and COMET. Figure \ref{en-vi-bleu-al}, \ref{en-vi-comet-al} and Table \ref{en-vi-num-results} report the results on WMT15 En$\rightarrow$Vi with Transformer-small, which also present the superior performance of SM$^2$-Uni and SM$^2$-Bi.

\subsection{Robustness of SM$^2$ to Sentence Length}
\begin{figure}
	\centering
	\small
        \subfigure[SM$^2$-Bi]{
            \includegraphics[width=0.45\textwidth]{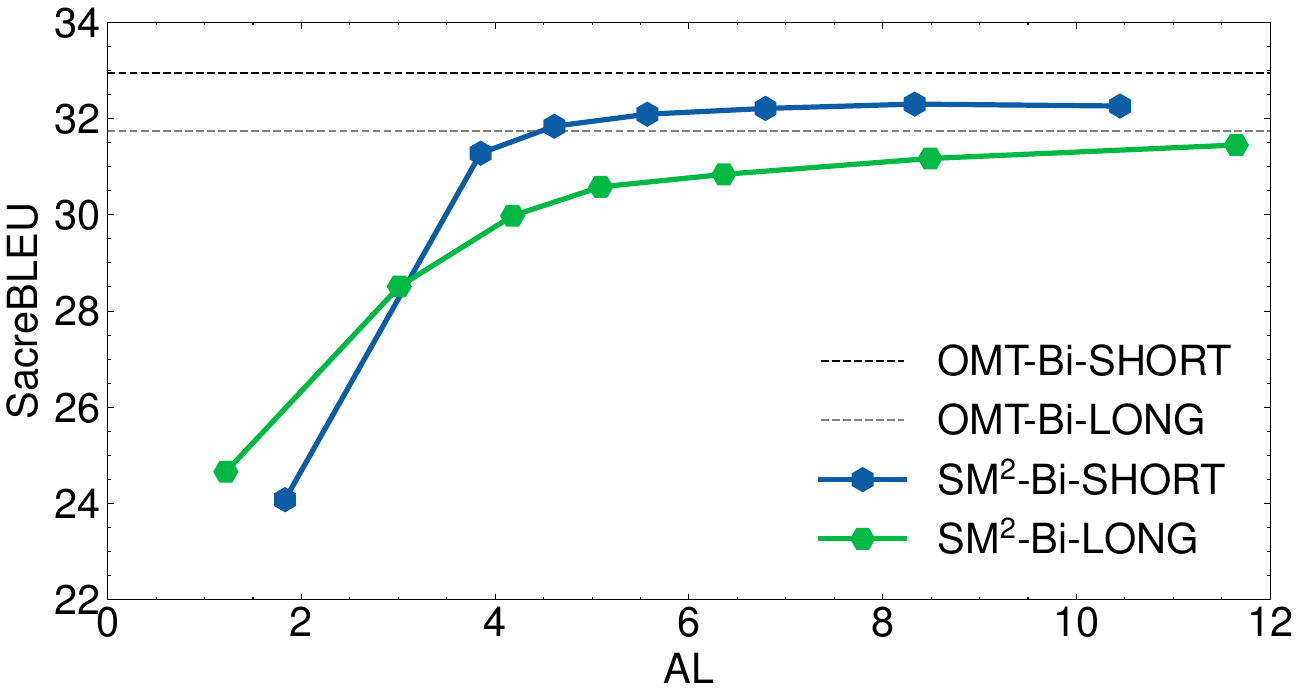} 
            \label{long-simt}
        }
        \subfigure[SM$^2$-Uni]{
            \includegraphics[width=0.45\textwidth]{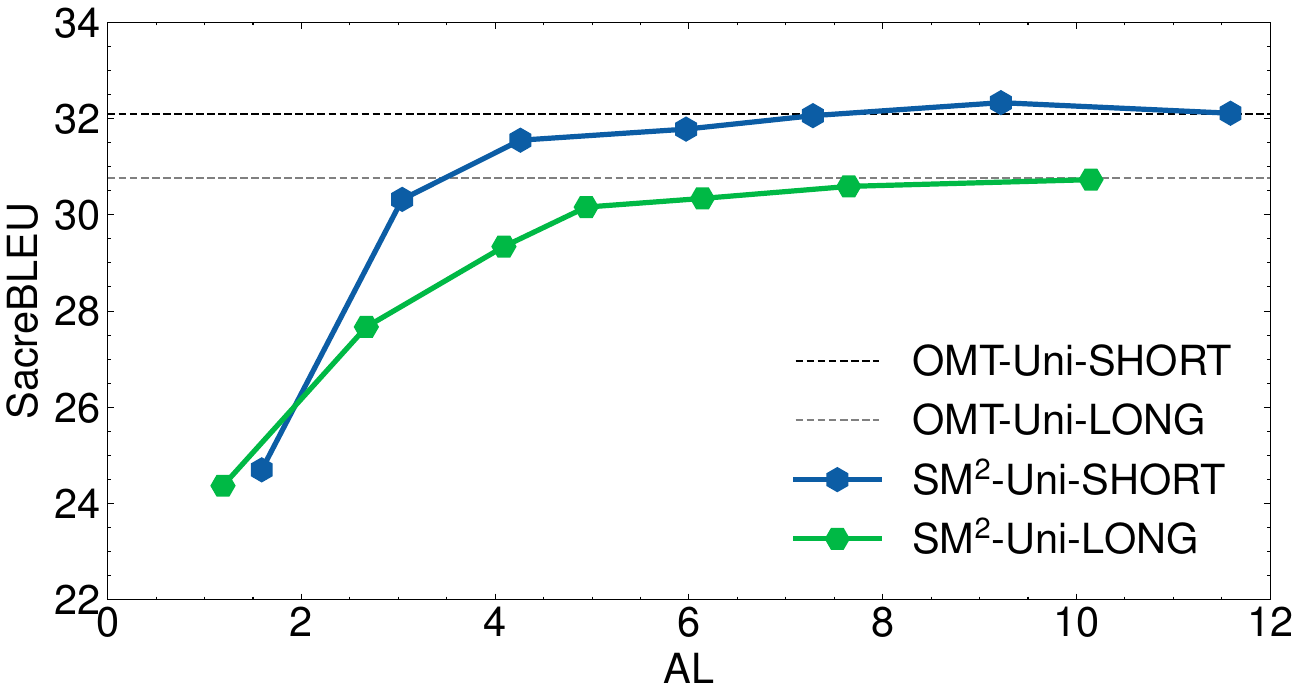}
            \label{short-simt}
        }
	\caption{Translation quality against latency of SM$^2$ on LONG and SHORT groups. We provide the performance of OMT models for a clearer comparison.}
	\label{long-short-simt}
    \end{figure}

To validate the robustness of SM$^2$ to Sentence Length, we conduct additional experiments on De$\rightarrow$En SiMT tasks. Specifically, we divide the test set into two groups based on sentence length: LONG group and SHORT group. The average lengths and the number of sentences in each group are shown in Table \ref{statics-length}. Then, we test SM$^2$-Bi and SM$^2$-Uni separately on these two groups. The translation quality under different latency levels for SM$^2$-Bi and SM$^2$-Uni are presented in Figure \ref{long-short-simt}. For clearer comparison, we also provide the performances of OMT models (OMT-Bi, OMT-Uni) on LONG and SHORT groups. 

The results in Figure \ref{long-short-simt} indicate that when applied to longer sentences, the performance changes of SM$^2$ are similar to OMT models in both unidirectional and bidirectional encoder settings. Since the performance of OMT models unavoidably drops as the sentences become longer \cite{neishi2019relation,kang2020dynamic,ma2020simple,varivs2021sequence,zhang2023layoutdit}, it is not SM$^2$ that triggers the decrease of translation quality. Therefore, SM$^2$ is still effective on long sentences.

\section{Effect of Prohibition on Policy}
\label{restrict-sa}
\begin{figure}
	\centering
	\small
        \includegraphics[width=0.45\textwidth]{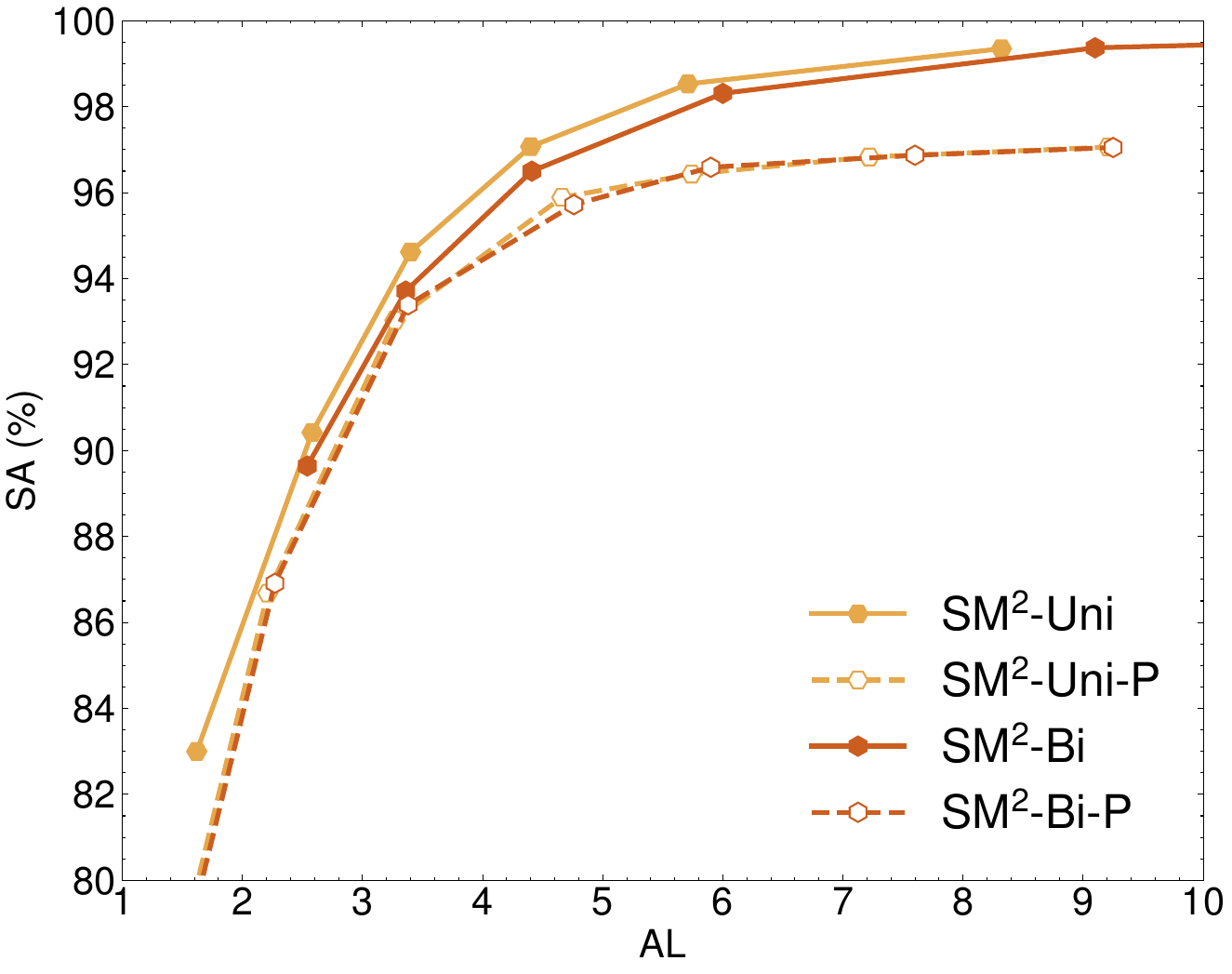}
	\caption{Evaluation of policies in SM$^2$ with and without prohibition. We calculate SA ($\uparrow$) under different latency levels.}
	\label{restrict-policy-eval}
    \end{figure}

To further validate that the prohibition of exploration negatively affects the policy. We compare the SA of SM$^2$ with and without the prohibition on RWTH dataset. The results in Figure \ref{restrict-policy-eval} indicate that the prohibition makes SM$^2$ learn a worse policy. Therefore, we can conclude that the prohibition will hurt the quality of policy. This further presents the advantage of SM$^2$ in sufficiently exploring all states through Prefix Sampling.

\begin{table}[]
\centering
\small
\begin{tabular}{cccc}
\toprule[1pt]
\multicolumn{4}{c}{\textbf{Chinese$\rightarrow$English}}           \\
\midrule
\multicolumn{4}{c}{wait-$k$}       \\
$k$      & AL    & SacreBLEU & COMET \\
1      & -0.60 & 23.14     & 67.06 \\
3      & 3.03  & 31.94     & 73.91 \\
5      & 4.96  & 35.56     & 75.87 \\
7      & 6.87  & 37.50     & 76.99 \\
9      & 8.82  & 38.90     & 77.85 \\
\midrule
\multicolumn{4}{c}{m-wait-$k$}     \\
$k$     & AL    & SacreBLEU & COMET \\
1      & 0.72  & 28.06     & 70.85 \\
3      & 2.80  & 32.41     & 74.29 \\
5      & 4.76  & 35.05     & 75.81 \\
7      & 6.81  & 36.68     & 76.86 \\
9      & 8.64  & 37.61     & 77.37 \\
\midrule
\multicolumn{4}{c}{HMT}            \\
$(L,K)$  & AL    & SacreBLEU & COMET \\
(2,4)  & 2.93  & 35.59     & 76.90 \\
(3,6)  & 4.52  & 37.81     & 78.08 \\
(5,6)  & 6.11  & 39.41     & 78.73 \\
(7,6)  & 7.69  & 40.33     & 79.11 \\
(9,8)  & 9.64  & 41.37     & 79.58 \\
(11,8) & 11.35 & 41.75     & 79.85 \\
\midrule
\multicolumn{4}{c}{ITST}           \\
$\delta$  & AL    & SacreBLEU & COMET \\
0.2    & 0.62  & 30.31     & 73.66 \\
0.3    & 2.88  & 35.87     & 77.02 \\
0.4    & 4.88  & 39.27     & 78.41 \\
0.5    & 6.94  & 41.20     & 79.27 \\
0.6    & 9.17  & 42.23     & 79.68 \\
0.7    & 11.40 & 42.75     & 79.93 \\
\midrule
\multicolumn{4}{c}{SM$^2$-Uni}        \\
$\gamma$  & AL    & SacreBLEU & COMET \\
0.3    & -0.63 & 29.52     & 73.62 \\
0.4    & 1.99  & 36.16     & 77.02 \\
0.5    & 4.56  & 39.94     & 78.66 \\
0.55   & 6.24  & 41.06     & 79.13 \\
0.6    & 8.51  & 42.21     & 79.50 \\
0.65   & 9.75  & 42.54     & 79.61 \\
\midrule
\multicolumn{4}{c}{SM$^2$-Bi}         \\
$\gamma$  & AL    & SacreBLEU & COMET \\
0.3    & -0.14 & 31.41     & 75.00 \\
0.4    & 2.35  & 37.77     & 78.09 \\
0.5    & 4.68  & 41.15     & 79.42 \\
0.55   & 6.19  & 42.47     & 79.91 \\
0.6    & 8.37  & 43.51     & 80.21 \\
0.65   & 11.61 & 44.34     & 80.45 \\
\bottomrule[1pt]
\end{tabular}
\caption{Numerical results on LDC Zh$\rightarrow$En. }
\label{zh-en-num-results}
\end{table}

\begin{table}[]
\centering
\small
\begin{tabular}{cccc}
\toprule[1pt]
\multicolumn{4}{c}{\textbf{German$\rightarrow$English}}             \\
\midrule
\multicolumn{4}{c}{wait-$k$}         \\
$k$      & AL    & SacreBLEU & COMET \\
1        & 0.10  & 20.11     & 70.74 \\
3        & 3.44  & 26.34     & 76.24 \\
5        & 6.00  & 28.96     & 78.44 \\
7        & 8.08  & 29.52     & 78.92 \\
9        & 9.86  & 30.23     & 79.71 \\
\midrule
\multicolumn{4}{c}{m-wait-$k$}       \\
$k$      & AL    & SacreBLEU & COMET \\
1        & 0.03  & 20.71     & 70.49 \\
3        & 2.94  & 24.85     & 74.49 \\
5        & 5.48  & 27.43     & 76.80 \\
7        & 7.66  & 28.2      & 77.67 \\
9        & 9.63  & 28.87     & 78.23 \\
\midrule
\multicolumn{4}{c}{HMT}              \\
$(L,K)$  & AL    & SacreBLEU & COMET \\
(2,4)    & 2.20  & 25.67     & 75.66 \\
(3,6)    & 3.58  & 28.29     & 77.94 \\
(5,6)    & 4.96  & 29.33     & 78.76 \\
(7,6)    & 6.58  & 29.47     & 79.23 \\
(9,8)    & 8.45  & 30.25     & 79.82 \\
(11,8)   & 10.18 & 30.29     & 79.74 \\
\midrule
\multicolumn{4}{c}{ITST}             \\
$\delta$ & AL    & SacreBLEU & COMET \\
0.2      & 2.27  & 25.17     & 75.17 \\
0.3      & 2.85  & 26.94     & 76.86 \\
0.4      & 3.83  & 28.58     & 77.98 \\
0.5      & 5.47  & 29.51     & 78.85 \\
0.6      & 7.60  & 30.46     & 79.28 \\
0.7      & 10.17 & 30.74     & 79.53 \\
0.8      & 12.72 & 30.84     & 79.61 \\
\midrule
\multicolumn{4}{c}{SM$^2$-Uni}       \\
$\gamma$ & AL    & SacreBLEU & COMET \\
0.3      & 1.39  & 24.68     & 75.58 \\
0.4      & 2.4   & 27.88     & 78.09 \\
0.5      & 3.56  & 29.6      & 79.51 \\
0.55     & 5.2   & 30.67     & 80.28 \\
0.6      & 6.33  & 30.86     & 80.36 \\
0.65     & 8.06  & 30.89     & 80.42 \\
0.7      & 10.74 & 31.08     & 80.53 \\
\midrule
\multicolumn{4}{c}{SM$^2$-Bi}        \\
$\gamma$ & AL    & SacreBLEU & COMET \\
0.3      & 1.52  & 24.74     & 75.96 \\
0.4      & 2.73  & 28.48     & 78.85 \\
0.5      & 3.73  & 30.17     & 80.21 \\
0.55     & 5.49  & 31.11     & 80.83 \\
0.6      & 7.03  & 31.42     & 81.00 \\
0.65     & 9.22  & 31.65     & 81.18 \\
0.7      & 12.33 & 31.92     & 81.25 \\
\bottomrule[1pt]
\end{tabular}
\caption{Numerical results on WMT15 De$\rightarrow$En. }
\label{de-en-num-results}
\end{table}

\begin{table}[]
\centering
\small
\begin{tabular}{cccc}
\toprule[1pt]
\multicolumn{4}{c}{\textbf{English$\rightarrow$Romanian}} \\
\midrule
\multicolumn{4}{c}{wait-$k$}                     \\
$k$         & AL       & SacreBLEU    & COMET    \\
1           & 2.70     & 26.62        & 74.12    \\
3           & 5.05     & 29.74        & 77.52    \\
5           & 7.18     & 31.61        & 78.54    \\
7           & 9.10     & 31.86        & 79.20    \\
9           & 10.92    & 31.89        & 78.97    \\
\midrule
\multicolumn{4}{c}{m-wait-$k$}                   \\
$k$         & AL       & SacreBLEU    & COMET    \\
1           & 2.66     & 26.65        & 74.27    \\
3           & 5.07     & 30.11        & 77.44    \\
5           & 7.18     & 31.05        & 78.35    \\
7           & 9.07     & 31.44        & 78.71    \\
9           & 10.89    & 31.37        & 78.62    \\
\midrule
\multicolumn{4}{c}{HMT}                          \\
$(L,K)$     & AL       & SacreBLEU    & COMET    \\
(1,2)       & 1.98     & 24.11        & 71.73    \\
(2,2)       & 2.77     & 27.18        & 74.85    \\
(4,2)       & 4.47     & 30.41        & 77.65    \\
(5,4)       & 5.48     & 31.56        & 78.80    \\
(6,4)       & 6.45     & 31.88        & 78.94    \\
(7,6)       & 7.41     & 31.85        & 79.17    \\
(9,6)       & 9.24     & 31.98        & 79.05    \\
\midrule
\multicolumn{4}{c}{ITST}                         \\
$\delta$    & AL       & SacreBLEU    & COMET    \\
0.1         & 2.75     & 22.76        & 71.19    \\
0.2         & 3.25     & 28.40        & 75.58    \\
0.3         & 5.09     & 30.52        & 77.53    \\
0.4         & 7.47     & 31.37        & 78.28    \\
0.45        & 8.81     & 31.62        & 78.49    \\
0.5         & 10.30    & 31.63        & 78.51    \\
0.55        & 11.69    & 31.74        & 78.73    \\
\midrule
\multicolumn{4}{c}{SM$^2$-Uni}                   \\
$\gamma$    & AL       & SacreBLEU    & COMET    \\
0.3         & 2.52     & 27.85        & 75.45    \\
0.4         & 2.72     & 29.21        & 76.62    \\
0.5         & 3.16     & 30.21        & 77.59    \\
0.6         & 4.17     & 31.20        & 78.26    \\
0.65        & 5.13     & 31.56        & 78.58    \\
0.7         & 6.56     & 31.72        & 78.77    \\
0.75        & 8.67     & 31.67        & 78.98    \\
\midrule
\multicolumn{4}{c}{SM$^2$-Bi}                    \\
$\gamma$    & AL       & SacreBLEU    & COMET    \\
0.3         & 2.60     & 28.74        & 76.81    \\
0.4         & 2.91     & 30.27        & 78.20    \\
0.5         & 3.57     & 31.33        & 79.04    \\
0.6         & 5.11     & 32.03        & 79.56    \\
0.65        & 6.51     & 32.40        & 79.90    \\
0.7         & 8.15     & 32.59        & 79.85    \\
0.75        & 10.10    & 32.74        & 79.95   \\
\bottomrule[1pt]
\end{tabular}
\caption{Numerical results on WMT16 En$\rightarrow$Ro. }
\label{en-ro-num-results}
\end{table}
\begin{table}[]
\centering
\small
\begin{tabular}{cccc}
\toprule[1pt]
\multicolumn{4}{c}{\textbf{English$\rightarrow$Vietnamese}} \\
\midrule
\multicolumn{4}{c}{wait-$k$}                     \\
$k$          & AL      & SacreBLEU    & COMET    \\
1            & 2.49    & 25.29        & 68.89    \\
3            & 4.28    & 28.03        & 70.28    \\
5            & 6.07    & 28.73        & 70.56    \\
7            & 7.89    & 28.72        & 70.72    \\
9            & 9.57    & 28.78        & 70.75    \\
\midrule
\multicolumn{4}{c}{m-wait-$k$}                   \\
$k$          & AL      & SacreBLEU    & COMET    \\
1            & 2.78    & 27.02        & 69.88    \\
3            & 4.38    & 28.59        & 70.61    \\
5            & 6.12    & 28.74        & 70.75    \\
7            & 7.88    & 28.69        & 70.78    \\
9            & 9.61    & 28.78        & 70.83    \\
\midrule
\multicolumn{4}{c}{HMT}                          \\
$(L,K)$      & AL      & SacreBLEU    & COMET    \\
(1,2)        & 2.9     & 27.69        & 70.22    \\
(4,2)        & 5.33    & 29.23        & 71.04    \\
(5,4)        & 6.23    & 29.36        & 71.01    \\
(6,4)        & 7.1     & 29.34        & 71.15    \\
(7,6)        & 8.01    & 29.42        & 70.99    \\
\midrule
\multicolumn{4}{c}{ITST}                         \\
$\delta$     & AL      & SacreBLEU    & COMET    \\
0.1          & 3.28    & 28.55        & 70.61    \\
0.15         & 4.52    & 29.04        & 70.90    \\
0.2          & 5.72    & 29.01        & 70.89    \\
0.25         & 8.38    & 29.13        & 70.83    \\
0.3          & 9.69    & 29.24        & 70.95    \\
\midrule
\multicolumn{4}{c}{SM$^2$-Uni}                   \\
$\gamma$     & AL      & SacreBLEU    & COMET    \\
0.6          & 1.87    & 28.29        & 70.46    \\
0.7          & 3.04    & 28.82        & 70.84    \\
0.75         & 5.31    & 29.28        & 71.16    \\
0.8          & 6.25    & 29.41        & 71.27    \\
0.9          & 9.45    & 29.57        & 71.30    \\
\midrule
\multicolumn{4}{c}{SM$^2$-Bi}                    \\
$\gamma$     & AL      & SacreBLEU    & COMET    \\
0.6          & 2.62    & 28.70         & 70.74    \\
0.7          & 4.47    & 29.38        & 71.27    \\
0.75         & 5.95    & 29.73        & 71.44    \\
0.8          & 7.07    & 29.75        & 71.40    \\
0.9          & 8.18    & 29.79        & 71.34   \\
\bottomrule[1pt]
\end{tabular}
\caption{Numerical results on WMT15 En$\rightarrow$Vi. }
\label{en-vi-num-results}
\end{table}

\begin{table}[]
\begin{tabular}{crr}
\toprule
\multicolumn{1}{l}{\textbf{}}              & \textbf{LONG}                & \textbf{SHORT}               \\
\midrule
{ Average Sentence Length}      & { 36.95} & { 14.07} \\
{ Number of Sentences} & { 1085}  & { 1084} \\
\bottomrule
\end{tabular}
\caption{Statistics on the average sentence length and number of sentences for LONG and SHORT groups.}
\label{statics-length}
\end{table}


\end{document}